\documentclass{article} % For LaTeX2e
\usepackage{iclr2024_conference,times}

\usepackage{amsmath,amsfonts,bm,amssymb}

\def\eqref#1{equation~\ref{#1}}
\def\1{\bm{1}}

\DeclareMathAlphabet{\mathsfit}{\encodingdefault}{\sfdefault}{m}{sl}
\SetMathAlphabet{\mathsfit}{bold}{\encodingdefault}{\sfdefault}{bx}{n}

\newcommand{\lr}{\alpha}

\usepackage{hyperref}
\usepackage{url}
\usepackage{caption}
\usepackage{subcaption}
\usepackage{graphicx}
\usepackage{cancel}
\usepackage{wrapfig}

\title{Neglected Hessian component explains mysteries in Sharpness regularization}

\author{
Yann Dauphin and Atish Agarwala\\
Google Deepmind \\
\texttt{\{ynd, thetish\}@google.com} \\
\And
Hossein Mobahi \\
Google Research \\
\texttt{hmobahi@google.com}
}

\usepackage{mathmacros}

\newcommand{\diag}{{\rm diag}}

\newcommand{\X}{\m{X}}

\newcommand{\W}{\m{W}}

\renewcommand{\L}{L}

\newcommand{\Hmat}{\m{H}}

\newcommand{\ntk}{\hat{\Theta}}

\newcommand{\Lo}{\mathcal{L}}

\renewcommand{\lr}{\eta}
\newcommand{\rad}{\rho}
\newcommand{\x}{\m{x}}
\newcommand{\y}{\m{y}}

\newcommand{\z}{\m{z}}
\renewcommand{\th}{\sm{\theta}}
\newcommand{\ep}{\sm{\epsilon}}
\newcommand{\J}{\m{J}}

\newcommand{\Am}{\m{A}}
\newcommand{\Bm}{\m{B}}

\newcommand{\h}{\m{h}}

\newcommand{\D}{D}

\newcommand{\relu}{{\rm ReLU}}
\newcommand{\GELU}{{\rm GELU}}

\newcommand{\gvec}{\m{g}}
\newcommand{\mname}{Nonlinear Modeling Error}
\newcommand{\mshort}{NME}
\newcommand{\gpen}[1]{\Lo_{pen,#1}}
\newcommand{\dad}{\mathcal{D}_{AD}}

\newif\ifcomments
\commentsfalse % uncomment to turn all colored comments off

\newif\ifrebuttal
\rebuttalfalse % uncomment to turn rebuttal coloring off

\ifcomments
\newcommand{\ynd}[1]{{\color{blue}[YD: #1]}}
\newcommand{\hmb}[1]{{\color{purple}[HM: #1]}}
\newcommand{\aga}[1]{{\color{red}[AA: #1]}}
\else
\newcommand{\ynd}[1]{}
\newcommand{\hmb}[1]{}
\newcommand{\aga}[1]{}
\fi

\ifrebuttal
\newcommand{\reb}[1]{{\color{magenta} #1}}
\else
\newcommand{\reb}[1]{#1}
\fi

\ifrebuttal
\newcommand{\arx}[1]{{\color{blue} #1}}
\else
\newcommand{\arx}[1]{#1}
\fi

\iclrfinalcopy % Uncomment for camera-ready version, but NOT for submission.
\begin{document}

\maketitle

\begin{abstract}
Recent work has shown that methods like SAM which either explicitly or implicitly penalize second order information can improve generalization in deep learning. 
% However this success is mixed; gradient norm penalties work on some architectures but fail on closely related ones, and seemingly similar methods like weight noise tend not to work at all.
Seemingly similar methods like weight noise and gradient penalties often fail to provide such benefits.
We show that these differences can be explained by the structure of the Hessian of the loss. First, we show that a common decomposition of the Hessian can be quantitatively interpreted as separating the feature exploitation from feature exploration. The feature exploration, which can be described by the \mname{} matrix (\mshort{}), is commonly neglected in the literature since it vanishes at interpolation. Our work shows that the \mshort{} is in fact important as it can explain why gradient penalties are sensitive to the choice of activation function. Using this insight we design interventions to improve performance.
We also provide evidence that challenges the long held equivalence of weight noise and gradient penalties. This equivalence relies on the assumption that the \mshort{} can be ignored, which we find does not hold for modern networks since they involve significant feature learning.
% We also provide evidence that the difference between weight noise and gradient penalties is related to the fact that weight noise penalizes the \mshort{} directly - which is important in the training dynamics of modern neural networks which involve significant feature learning.
We find that regularizing feature exploitation but not feature exploration yields performance similar to gradient penalties.

% Recent work has shown that first order methods like SAM which implicitly penalize second order information  can improve generalization in deep learning. Seemingly similar methods like weight noise and gradient penalties often fail to provide such benefits. We show that these differences can be explained by the structure of the Hessian of the loss. First, we show that a common decomposition of the Hessian can be quantitatively interpreted as separating the feature exploitation from feature exploration. The feature exploration, which can be described by the \mname{} matrix (\mshort{}), is commonly neglected in the literature since it vanishes at interpolation. Our work shows that the \mshort{} is in fact important as it can explain why gradient penalties underperform for certain architectures. Furthermore, we provide evidence that challenges the long held equivalence of weight noise and gradient penalties. This equivalence relies on the assumption that the \mshort{} can be ignored, which we find does not hold for modern networks since they involve significant feature learning. Intriguingly, we find that regularizing feature exploitation but not feature exploration yields performance comparable to SAM. 
% This suggests that properly controlling regularization on the two parts of the Hessian is important for the success of many second order methods.
\end{abstract}

\section{Introduction\aga{COMMENTS ON}}

% \arx{
% TODO:
% \begin{enumerate}
%     \item Put focus instead on showing existence of cases where NME cannot be neglected. Previously, focus was on the mysteries but the common thread is actually "this is a case where NME affects training".
%     \item Say "methods that use second order information" instead of "second order methods".
%     \item The problem with second mystery is that we were trying to do at least two things but not disentangling them: 1) show NME is important for weight noise 2) explain the difference between penalty sam and weight noise. We will move point 2 to discussion instead and make that section fully focused on just showing that NME is important.
%     \item Move Section 2.1 and merge it with 4.4
%     \item Replace "penalty sam" with "gradient penalty" in main text.
%     \item Fix all section and figure references
% \end{enumerate}
% }

There is a long history in machine learning of trying to use information about
the loss landscape geometry to improve gradient-based learning. This has ranged from attempts to use the Fisher information matrix to improve optimization \citep{martens_optimizing_2015},
to trying to regularize the Hessian to improve generalization \citep{sankar2021deeper}.
More recently, first order methods which implicitly use or penalize second order
quantities have been used successfully,
including the \emph{sharpness aware minimization}
(\texttt{SAM}) algorithm \citep{foret2020sharpness}. On the other hand, there are many approaches to use second order information
which once seemed promising but have had limited success \citep{dean2012large}. These include
methods like weight noise \citep{an1996effects} and gradient norm penalties, which have shown mixed success.

Part of the difficulty of using second order information is the difficulty of
working with the Hessian of the loss. With the large number of parameters in deep learning
architectures, as well as the large number of datapoints, many algorithms use
stochastic methods to approximate statistics of the Hessian
\cite{martens_optimizing_2015, liu2023sophia}. However,
there is a \emph{conceptual} difficulty as well which arises from the
complicated structure of the Hessian itself. Methods often involves
approximating the Hessian via the Gauss-Newton (GN) matrix - which is
PSD for convex losses.
 \reb{This is beneficial
for conditioners which try to maintain monotonicity of gradient flow via a PSD
transformation. Thus indefinite part of the Hessian is often neglected due to
its complexity.}

In this work we show that it is important to consider \emph{both} parts of the
Hessian \arx{to understand certain methods that use second order information for regularization}. We show that
\reb{with saturating non-linearities,}
the GN part of the Hessian is related to \emph{exploiting} existing linear
structure, while the indefinite part of the Hessian, which we dub the
\emph{\mname~matrix} (\mshort), is related to \emph{exploring} the effects of
switching to different \reb{multi-}linear regions.
% Removed because it is redundant with next bullet points
% We show that the \mshort{} depends
% heavily on the second derivative of the activation function. This suggests that
% \arx{methods that use second order information} may be \reb{more sensitive to details of the activation
% function compared to first order methods.}
\arx{In contrast to commonly held assumptions, this work identifies two distinct cases where neglecting the influence of the indefinite component of the Hessian is demonstrably detrimental:

\begin{itemize}
\item \textbf{Training with Gradient Penalties.} Our theoretical analysis reveals that the activation function controls the sparsity of information encoded within the indefinite component of the Hessian. Notably, we demonstrate that manipulating this sparsity by changing the activation function can transform previously ineffective gradient penalties into potent tools for improved generalization. To the best of our knowledge, this work is the first to show that methods using second order information are more sensitive to the choice of activation function.
% \item \textbf{Training with Weight Noise.} Conventional analysis of weight noise overlooks the indefinite component of the Hessian, but our experimental ablations demonstrate that it exerts a significant influence on generalization performance.

\item \textbf{Training with
Hessian penalties.} Conventional analysis of
weight noise casts it as a penalty on the GN part of the
Hessian, but in reality it also penalizes the NME. Our
experimental ablations show that the NME exerts a significant
influence on generalization performance.
\end{itemize}

}

% We then use our theoretical insights to guide experiments which show the
% following:
% \begin{itemize}
% \item We explain the inconsistencies between the success of \texttt{SAM} and
% failure of gradient penalty regularizers in certain architectures to the
% choice of activation functions, and rescue the performance of the gradient penalty
% by switching ReLU to GELU. To our knowledge we are the first to show that methods using second order information are more sensitive to the choice of activation function.
% \item We show that weight noise does not perform as well as the gradient penalty it is thought to approximate. We provide evidence that this is due to the analysis neglecting the important effect of the \mshort{} matrix, which weight noise implicitly penalizes.
% \item Furthermore, we show that penalizing the GN part of the Hessian
% directly while ignoring the \mname{} does seem to improve generalization.
% \end{itemize}

We conclude with a discussion about how these insights might be used to
design activation functions not with an eye towards forward or
backwards passes \citep{pennington_resurrecting_2017, martens_rapid_2021},
but for compatibility with \arx{methods that use second order information}.

\section{Understanding the structure of the Hessian}\label{sec:structure}

% The key hypothesis of this paper is that the structure of the Hessian can be used to explain the empirical phenomena of Sections \ref{sec:penalty_sam} and \ref{sec:weight_noise}.
\reb{In this section, we lay the theoretical ground work
for our experiments
by explaining the structure of the Hessian.} Given a model $\z(\th, \x)$ defined on parameters $\th$ and input $\x$,
and a loss function $\Lo(\z, \y)$ on the model outputs and labels $\y$, we can write the gradient of the training loss with respect to $\th$ as
\begin{equation}
\nabla_{\th}\Lo = \J^{\tpose}(\nabla_{\z}\Lo)
\end{equation}
where the Jacobian $\J \equiv \nabla_{\th}\z$.
The Hessian $\nabla_{\th}^2\Lo$ can be decomposed as:
\begin{equation}
\nabla_{\th}^2\Lo = \underbrace{\J^{\tpose} \Hmat_{\z} \J}_{\mathrm{GN}} + \underbrace{\nabla_{\z}\Lo \cdot \nabla^2_{\th} \z}_{\mathrm{NME}} 
\label{eqn:hess_decomp}
\end{equation}
where $\Hmat_{\z} \equiv \nabla^2_{\z}\Lo$.
The first term is often called the Gauss-Newton (GN) part of the Hessian
\citep{jacot_asymptotic_2020, martens_new_2020}.
If the loss function is convex with respect to the model outputs/logits
(such as for MSE and CE losses), then the GN matrix is positive semi-definite.
This term often contributes large eigenvalues. \reb{The second term has previously
been studied theoretically where it is called the
\emph{functional Hessian} \citep{singh_analytic_2021, singh_hessian_2023};
in order to avoid confusion with the overall Hessian we call it the \emph{\mname{}} matrix (\mshort{}).} It is in general 
indefinite and vanishes to zero at an interpolating minimum $\th^*$
where the model ``fits''the data ($\nabla_{z}\Lo(\th^*) = \boldsymbol{0}$), as can happen in
overparameterized settings.
Due to this, it is quite common for studies to drop this term 
entirely when dealing with the Hessian. For example, many second order optimizers 
approximate the Hessian $\nabla_{\th}^2\Lo$ with only the Gauss-Newton term 
\citep{martens2011learning,liu2023sophia}. It is also common  to neglect this term in 
theoretical analysis of the Hessian $\nabla_{\th}^2\Lo$ \citep{bishop1995training, sagun2017empirical}. However, we will 
show why this term should not be ignored.

\begin{figure}[h]
\centering
\begin{tabular}{cc}
\includegraphics[trim={0 2cm 2cm 3cm}, height=0.34\textwidth]{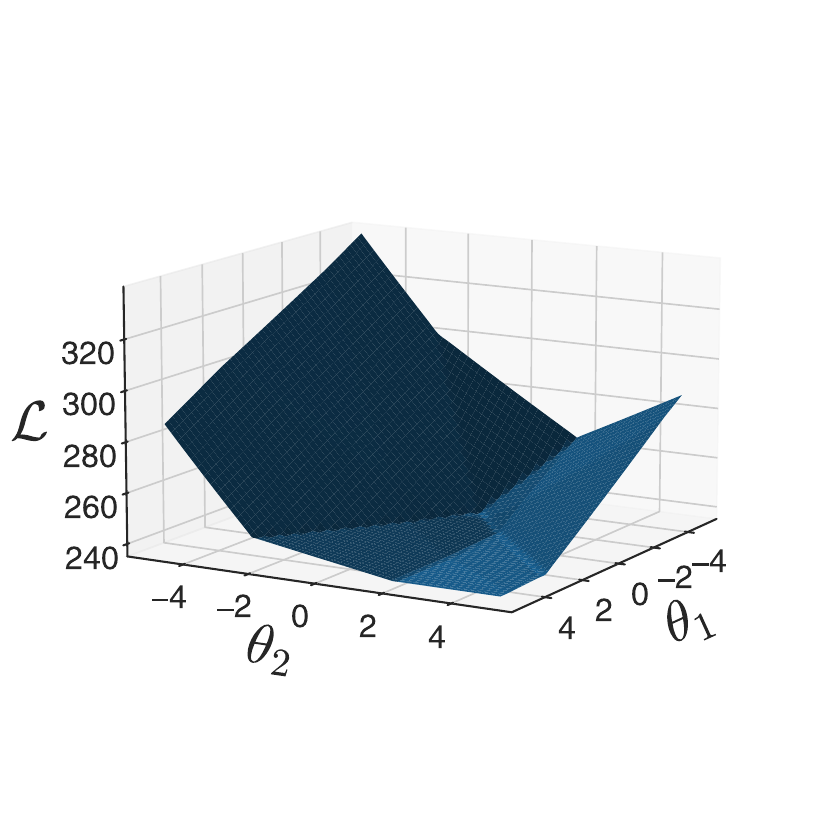}     & \includegraphics[height=0.34\textwidth]{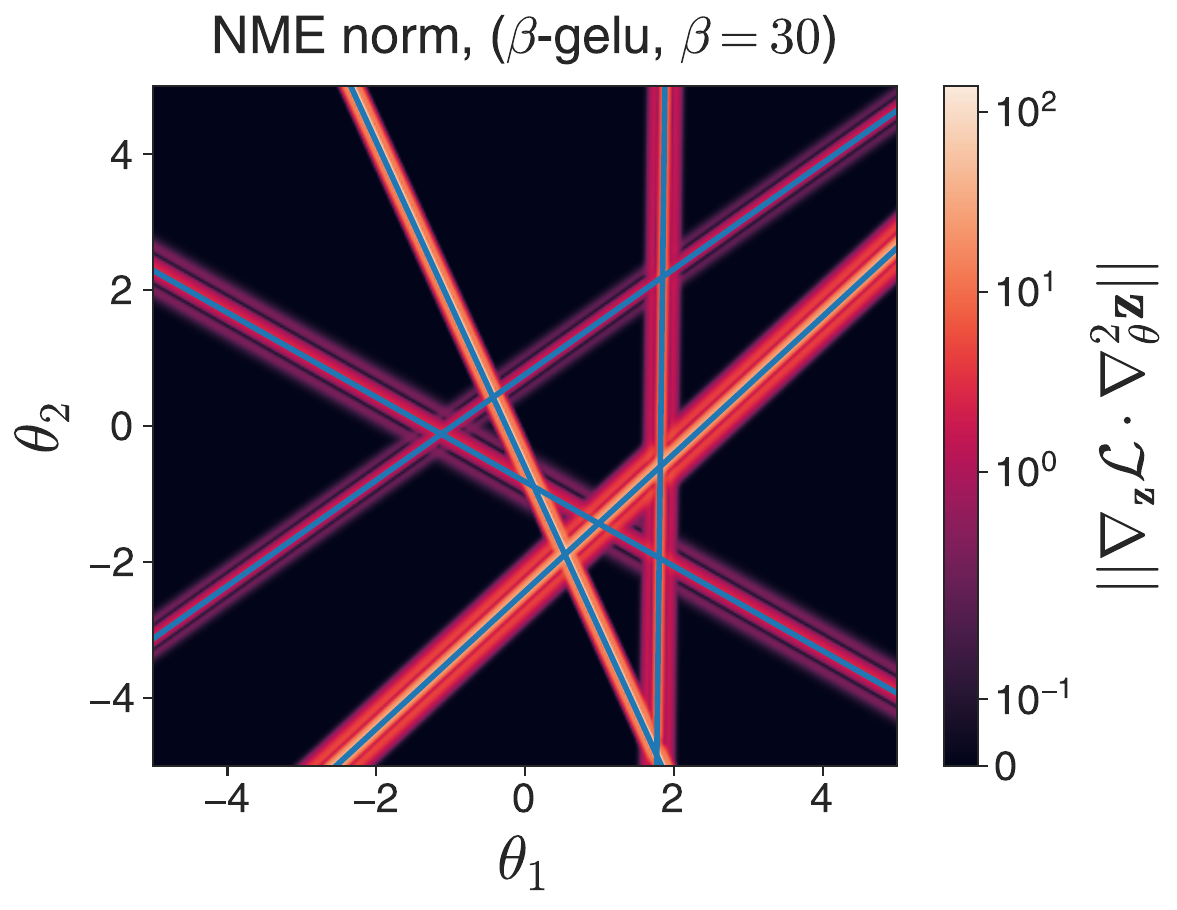}
\end{tabular}

\caption{Loss (left) and \mname{} matrix (\mshort{}) norm (right) as a function of
$2$ parameters in the same hidden layer of an
MLP (MSE loss, one datapoint). For ReLU activation model is piecewise multilinear,
and piecewise linear for parameters in same layer. Loss is piecewise quadratic
for parameters in same layer
(left). There is little \mshort{} information
accessible pointwise and the main features are the boundaries of the
piecewise linear regions (blue, right). For $\beta$-GELU, \mshort{} magnitude is high
only within distance $1/\beta$ of those boundaries. Therefore the \mshort{}
encodes information about the utility of switching between piecewise
multilinear regions.}
\label{fig:lin_regions_curvature}
\end{figure}

While the \mshort{} term can become small late in training,
it encodes significant information during training. More precisely, {\it it is the
only part of
Hessian that contains second order information from the model features} 
$\nabla_{\th}^{2}\z$. The GN matrix only contains 
second order information about the loss w.r.t. the logits with the term $\Hmat_{\z}$. 
All the information about the model function in the GN matrix is first-order. In 
fact, the GN matrix can be seen as the Hessian of an approximation of the loss 
where a first-order approximation of the model $\z(\th', \x) \approx \z(\th, \x) + \J \sm{\delta}$ ($\sm{\delta}=\th'-\th$) is used \citep{martens2011learning}
\begin{equation}
    \nabla^{2}_{\sm{\delta}}\Lo(\z(\theta, \x) + \J \sm{\delta}, \y) |_{\th'=\th} = \J^{\tpose} \Hmat_{\z} \J
\end{equation}
Thus we can see the GN matrix as the result of a linearization of the model and 
the \mshort{} as the part that takes into account the non-linear part of the 
model. The GN matrix exactly determines the
linearized (NTK) dynamics of training, and therefore controls
learning over small parameter changes when the features can be approximated as fixed
(see Appendix \ref{app:ntk_gn}). In contrast, the \mshort{} encodes information
about the \emph{changes} in the NTK \citep{agarwala_secondorder_2022}. For example
given a piecewise multilinear model like a ReLU network,
we can
think of the GN part of the Hessian as \emph{exploiting} the linear (NTK) structure, while the \mshort{} gives information on \emph{exploration} -
namely, the benefits of switching to a different multilinear region where different neurons are active. See Figure \ref{fig:lin_regions_curvature} for an illustration of this with ReLU model. We discuss this aspect further in Section \ref{sec:relu_gelu_deriv}.

The GN part may \emph{seem} like it must contain this second order
information due to its equivalence to the Fisher information matrix for losses that can
be written as negative log-likelihoods, like MSE and cross-entropy. For these, the Fisher information itself can be written as
the Hessian of a slightly different loss \citep{pascanu2013revisiting}:
\begin{equation}
    {\bf F} = \expect_{\hat{\bf y} \sim \m{p}_\z}\left[\nabla^{2}_{\th}\Lo(\z, \hat{\y})\right]
\end{equation}
where the only difference is that the labels $\hat{\y}$ are sampled from the model 
instead of the true labels. However, the \mshort{} is $0$ for this loss. For
example, in the case of MSE using Equation \ref{eqn:hess_decomp} we have
\begin{align}
    \expect_{\hat{\y} \sim \m{p}_\z}\left[\nabla^{2}_{\th} \Lo(\z, \hat{\y})\right] &= \expect_{\hat{\y} \sim \mathcal{N}(\z, \m{I})}\left[\J^{\tpose} \Hmat_{\z} \J + \nabla_{\z}\Lo(\z,\hat{\y}) \cdot \nabla^{2}_{\th}\z\right] \\
\label{eq:fisher_nme}
    &= \J^{\tpose} \Hmat_{\z} \J + \cancel{\expect_{\hat{\y} \sim \mathcal{N}(\z, \m{I})}[\z-\hat{\y}]} \cdot \nabla^{2}_{\th}\z
\end{align}
The second term in Equation \ref{eq:fisher_nme} (\mshort{}) vanishes because we are at the global minimum for this loss.

\section{Experimental Setup}

Our analysis of the Hessian begs an immediate question: when does the \mshort{}
affect learning algorithms? We conducted experimental studies to answer this question
in the context of curvature regularization algorithms which seek to promote
convergence to flat areas of the loss landscape. We use the following two setups
for the remainder of the paper:

{\bf Imagenet} We conduct experiments on the popular Imagenet dataset \citep{deng2009imagenet}. All experiments use the Resnet-50 architecture with the same setup and hyper-parameters as \cite{goyal2018accurate}, except that we use cosine learning rate decay \citep{loshchilov2016sgdr} over 300 epochs.

{\bf CIFAR-10} We also provide results on the CIFAR-10 dataset \citep{krizhevsky2009learning}.
All experiments use the \arx{WideResnet 28-10} architecture with the same setup and hyper-parameters as \cite{zagoruyko2016wide}, except for the use of cosine learning rate decay.

\section{
\arx{How \mshort{} affects training with gradient penalties}}
\label{sec:grad_pen}

%\section{When does Penalty \texttt{SAM} fail?}\label{sec:penalty_sam}

\arx{

% TODO:
% \begin{enumerate}
%     \item Move SAM stuff to discussion
%     \item Remove Original SAM from plots.
%     \item Add results for Augmented and Diminished ReLU
% \end{enumerate}

In this section we will show that the information contained in the \mshort{} has a critical impact on the effectiveness of gradient penalties for generalization. We define a gradient
penalty as an additive regularizer of the form:
\begin{equation}\label{eqn:gradient_penalty}
\gpen{p} = \rad||\nabla\Lo_{0}||^{p}
\end{equation}
for a base loss $\Lo_{0}$.
Gradient penalties have recently gained popularity as regularizers \citep{BarrettD21, smith2021origin, du2022sharpnessaware, zhao2022penalizing, reizinger2023samba};
this is in part due to their ability to reduce sharpness. In fact,
$\gpen{p}$ is closely related to Sharpness Aware Minimization (\texttt{SAM})
\citep{foret2020sharpness}. $p = 1$ corresponds to the original normalized formulation,
while $p = 2$ corresponds to the unnormalized formulation which is equally
effective and easier to analyze \citep{andriushchenko2022towards, agarwala_sam_2023}.
A more detailed description of the link between \texttt{SAM} and gradient penalties
can be found in Appendix \ref{app:sam_and_gpen}. We will focus on the $p = 1$ case in the remainder
of this section.

}

\arx{

\subsection{Gradient penalty update rules}
}
Consider the SGD update rule for the $p = 1$ gradient penalty and
base loss $\Lo_{0}$. With learning rate $\lr$, the parameters $\th$ evolve as:
\begin{equation}
\th_{t+1}-\th_{t} = -\lr\left(\nabla_{\th}\Lo_{0}+\frac{1}{||\nabla_{\th}\Lo_{0}||}\Hmat\nabla_{\th}\Lo_{0}\right),~\Hmat \equiv \nabla_{\th}^{2}\Lo_{0}
\label{eq:gpen_update}
\end{equation}
The additional contribution to the dynamics comes in the form of a Hessian-gradient
product.

\begin{figure}[ht]
     \centering
     \begin{subfigure}[b]{0.45\textwidth}
         \centering
         \includegraphics[width=\textwidth]{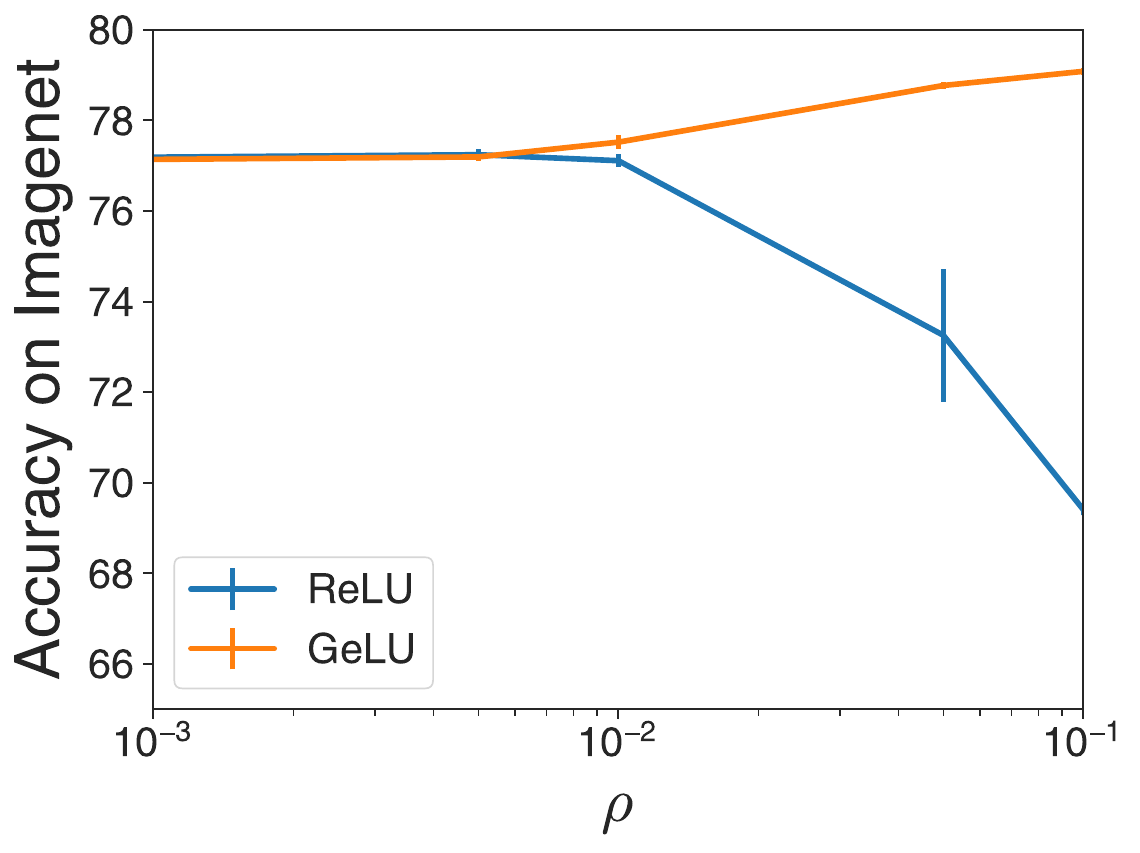}
         \caption{Imagenet}
         \label{fig:y equals x}
     \end{subfigure}
     \hfill
     \begin{subfigure}[b]{0.45\textwidth}
         \centering
         \includegraphics[width=\textwidth]{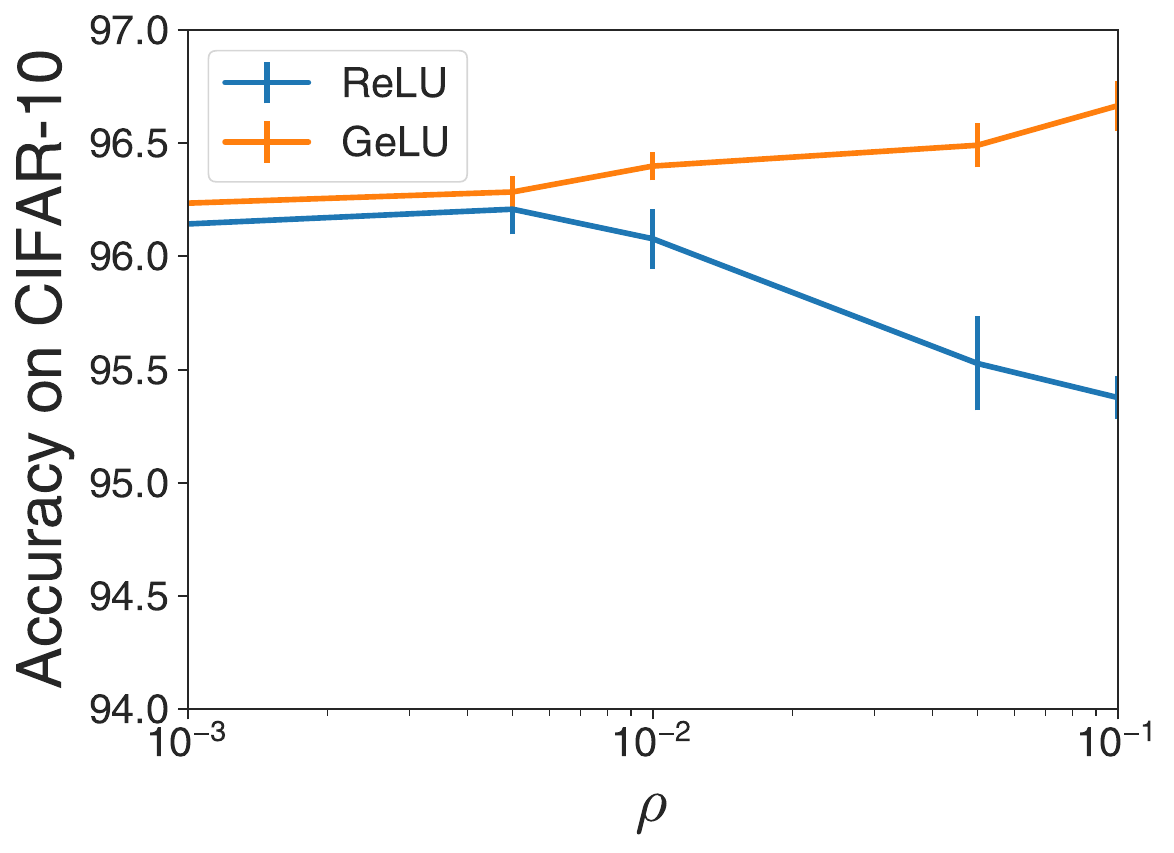}
         \caption{CIFAR-10}
         \label{fig:three sin x}
     \end{subfigure}
     % \begin{subfigure}[b]{0.45\textwidth}
     %     \centering
     %     \includegraphics[width=\textwidth]{images/imagenet_GELU_rho.pdf}
     %     \caption{Imagenet with GELU}
     %     \label{fig:y equals x}
     % \end{subfigure}
     % \hfill
     % \begin{subfigure}[b]{0.45\textwidth}
     %     \centering
     %     \includegraphics[width=\textwidth]{images/cifar10_GELU_rho.pdf}
     %     \caption{CIFAR-10 with GELU}
     %     \label{fig:three sin x}
     % \end{subfigure}
        \caption{Test Accuracy vs. $\rad$ for ReLU and GELU networks trained
        with gradient penalty ($p=1$, averaged over $5$ seeds). In both cases performance
        is similar without regularization but with regularization test accuracy increases
        for GELU until $\rad = 0.1$ and decreases for ReLU over a similar
        range.}
        \label{fig:activations_rho}
\end{figure}

Since the update rule explicitly involves the Hessian, 
a natural question is: do the GN and \mshort{} both play a significant role in the dynamics?
Or does the conventional wisdom hold - that the GN dominates? We explore this question
by starting with a simple experiment. In our Imagenet and CIFAR10 setups, we consider
networks trained with ReLU activations and networks trained with GELU activations.
Without regularization, both activation functions achieve similar test accuracy
($76.8$ for both on Imagenet). However, as the gradient penalty regularizer is added,
differences emerge with increasing $\rad$ (Figure \ref{fig:activations_rho}).
The performance of GELU networks increases with $\rad$ as high as $0.1$;
in contrast, the performance of ReLU networks is decreasing in $\rad$.

Even though both activations have similar forward passes,
the addition of the Hessian-gradient product seems to dramatically change the
learning dynamics. Since training without regularization seems to be similar across
the activation functions, we focus on the difference between the Hessians induced
by ReLU and GELU. As we will see, it is in fact the \mshort{} which is dramatically
different between the two activations.

\arx{
\subsection{Effect of Activation functions on the \mshort{}}
}

One important feature of the \mshort{} is that it depends on the \emph{second derivative}
of the activation function. We can demonstrate this most easily on a fully-connected
network, but the general principle applies to most common architectures.
Given an activation function $\phi$, a feedforward network
with $L$ layers on an input $\x_{0}$ defined iteratively by
\begin{equation}
\h_{\l} = \W_{\l}\x_{\l},~\x_{\l+1} = \phi(\h_{\l})
\end{equation}
The gradient of the model output $\x_{L}$ with respect to a weight matrix
$\W_{l}$ is given by
\begin{equation}
\frac{\partial \x_{L}}{\partial \W_{\l}} = \J_{L(\l+1)}\circ\phi'(\h_{l}) \otimes\x_{l}, ~ \J_{\l'\l} \equiv \prod_{m=\l}^{\l'-1}\phi'(\h_{m})\circ\W_{m}
\end{equation}
where $\circ$ is the Hadamard (elementwise) product.
The second derivative can be written as:
\begin{equation}\label{eqn:hessian_activation}
\frac{\partial^2 \x_{L}}{\partial \W_{\l}\partial \W_{m}} = \left[\frac{\partial\J_{L(\l+1)}}{\partial \W_{m}}\circ\phi'(\h_{l})+\J_{L(\l+1)}\circ\frac{\partial\phi'(\h_{l})}{\partial\W_{m}}\right] \otimes\x_{l}
\end{equation}
where without loss of generality $m\geq l$. The full analysis of this derivative
can be found in Appendix \ref{app:second_deriv}.
The key feature is that the majority of the terms have a factor of the
form
\begin{equation}
\frac{\partial \phi'(\h_{o})}{\partial\W_{m}} = \phi''(\h_{o})\circ \frac{\partial \h_{o}}{\partial\W_{m}}
\end{equation}
via the product rule - a dependence on $\phi''$.
On the diagonal $m = l$, all the terms depend on $\phi''$.
We note that a similar analysis can be found in Section 8.1.2 of
\cite{martens_new_2020}.

\arx{
\subsection{ReLU vs. GELU second derivatives}
}

\label{sec:relu_gelu_deriv}

The second derivative of the activation function is
key to controlling the statistics of the \mshort{}. Due to the
popularity of first order optimizers, activation functions have been designed to have 
well behaved
first derivatives - with little concern for second derivatives. Consider ReLU: it became popular
as a way to deal with gradient propagation issues from activations like $\tanh$;
however, it suffers from a ``missing curvature'' phenomenology -
mathematically, the ReLU second derivative is $0$ everywhere except the origin, where it is
undefined. In practical implementations it is set to $0$ at the origin as
well.
This implies that the diagonal of the \mshort{} is $0$ for ReLU in practice.

In contrast, GELU has a well-posed second derivative - and therefore a non-trivial
\mshort{}. We can study the difference between the GELU and ReLU by using the
$\beta$-GELU which interpolates between the two. It is given by
\begin{equation}
\beta\text{-GELU}(x) = x\Phi(\beta x)
\end{equation}
where $\Phi$ is the standard Gaussian CDF. We can recover GELU by setting $\beta=1$, and ReLU is recovered in the limit $\beta\to\infty$ (except for the second derivative
at the origin which as we will see it is undefined).
The second derivative is given by
\begin{equation}
\frac{d^2}{dx^2}\beta\text{-GELU}(x) = \frac{1}{\sqrt{2\pi \beta^{-2}}}e^{-x^2/2\beta^{-2}}\left[2-(x/\beta^{-1})^2\right]
\label{eq:2nd_deriv_GELU}
\end{equation}
For large $\beta$, this function is exponentially small when $x\gg \beta^{-1}$,
and $O(\beta)$ when $|x| = O(\beta^{-1})$. As $\beta$ increases the non-zero 
region becomes smaller while the non-zero value becomes larger such that
the integral is always $1$. This suggests that rather than being
uniformly $0$, the ReLU second derivative
is better described by Dirac delta ``function'' (really a distribution) -
$0$ except at the origin, where it is undefined, but still integrable to $1$.
Note that $\beta$-GELU for large $\beta$ is different from standard ReLU implementations
at the origin, since it has second derivative $\beta$, and not $0$, at the origin.

The choice of $\beta$ determines how much information the \mshort{}
can convey in a practical setting. This second derivative is large only when
the input to the activation
is within distance $1/\beta$ of $0$. In a deep network this corresponds to
being near the boundary of the piecewise multilinear regions where the activations
switch on and off. We can illustrate this using two parameters of an MLP
in the same layer, where the model is in fact piecewise linear with respect to
those parameters
(Figure \ref{fig:lin_regions_curvature}). The second derivative
serves as an ``edge detector''\footnote{In fact, the negative of the second order derivative of GELU is closely related to the Laplacian of Gaussian, which is a well-known edge-detector in image processing and computer vision.} (more generally, hyperplane detector),
and the \mshort{} can be used to probe the
usefulness of crossing these edges.

From Equation \ref{eqn:hessian_activation}, this means that for intermediate
$\beta$ many terms of the diagonal of the
\mshort{} will be non-zero at a typical point. However as $\beta$ increases,
the probability of terms being non-zero becomes low, but when they are non-zero
they are large - giving a sparse, spiky structure to the \mshort{}, especially
on the diagonal.
This leads to the \mshort{} becoming a high-variance estimator of local structure.
Therefore any methods seeking to use this information
explicitly are doomed to fail.

\begin{figure}[ht!]
     \centering
     \begin{subfigure}[b]{0.45\textwidth}
         \centering
         \includegraphics[width=\textwidth]{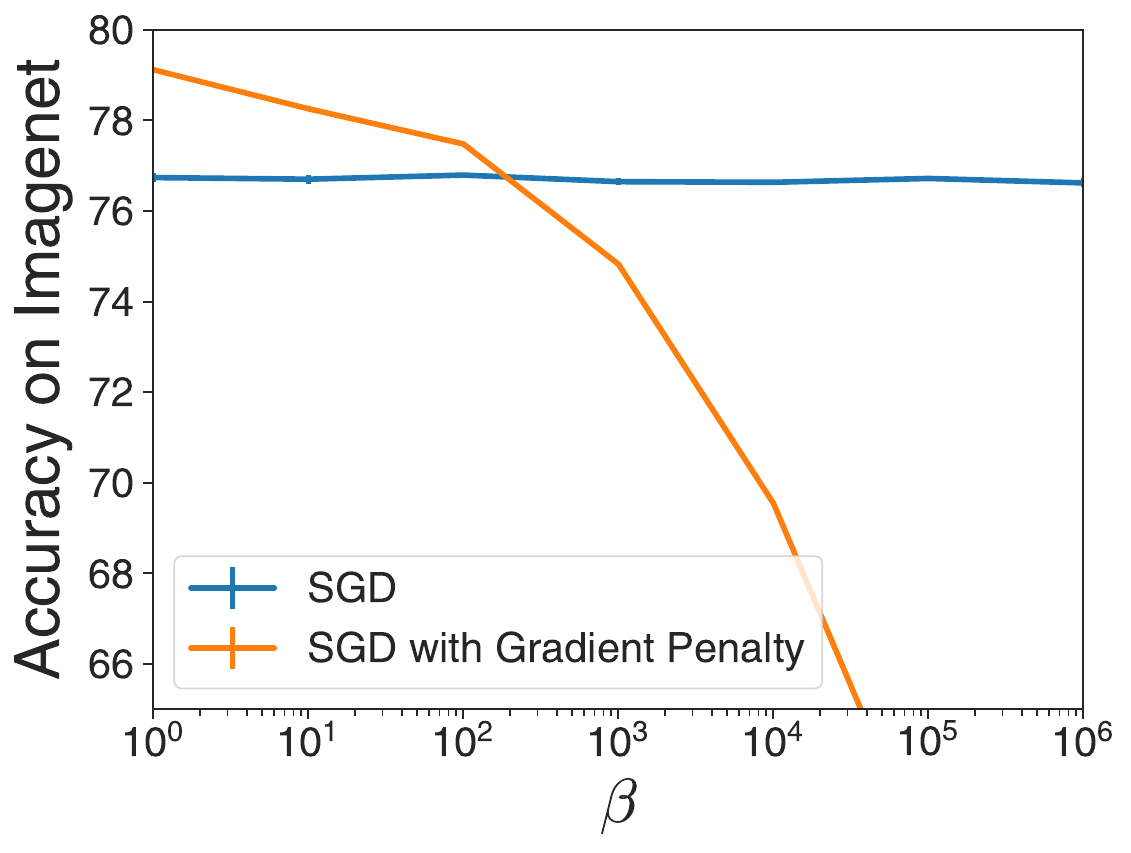}
         \caption{Imagenet}
         \label{fig:y equals x}
     \end{subfigure}
     \hfill
     \begin{subfigure}[b]{0.45\textwidth}
         \centering
         \includegraphics[width=\textwidth]{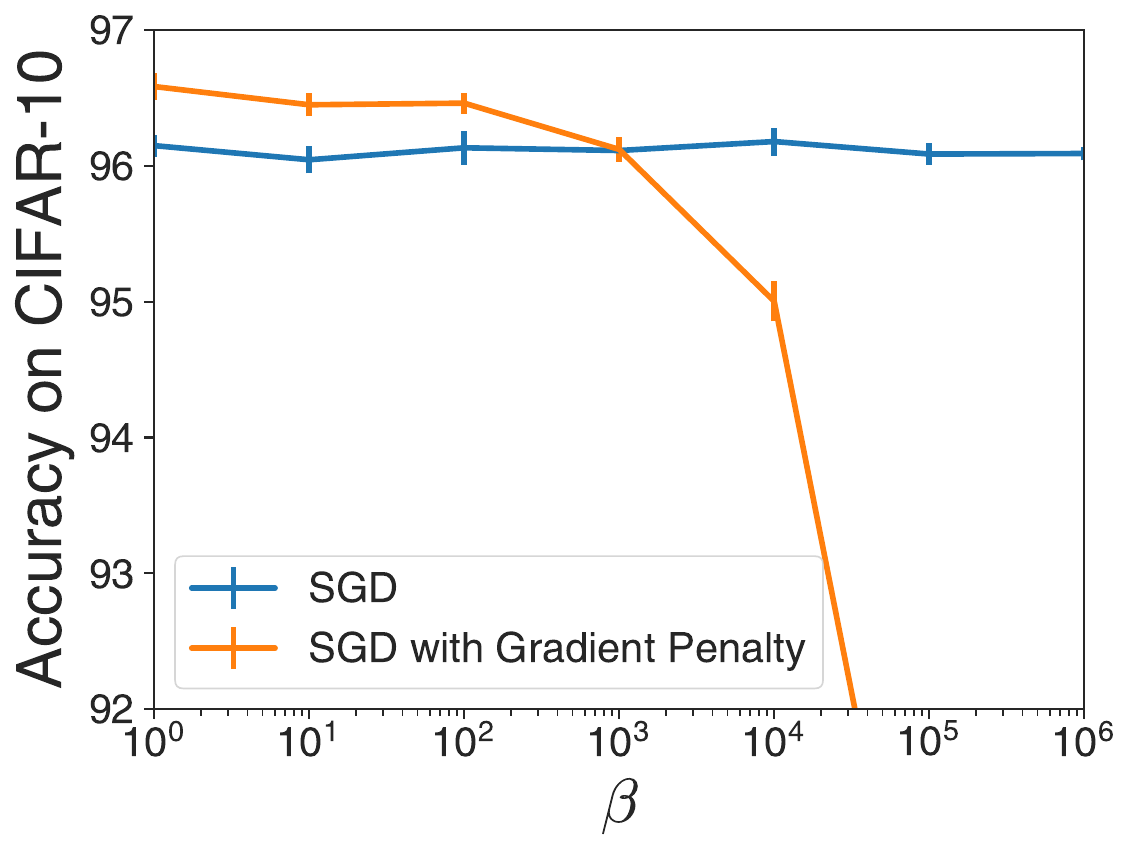}
         \caption{CIFAR-10}
         \label{fig:three sin x}
     \end{subfigure}
        \caption{Accuracy vs $\beta$ for SGD and SGD with gradient penalty ($\rho=0.1$) using $\beta$-GELU activations (average of $5$ seeds). \arx{We observe that accuracy decreases with larger $\beta$ with the gradient penalty but not without it}. \arx{As our theory suggests that the sparsity of the \mshort{} increases with $\beta$, this is evidence that it has significant impact on gradient penalties.}
        % We can see that as the $\beta$-GELU starts to approximates the ReLU accuracy decreases for $\rad>0$.
        }
        \label{fig:accuracy_beta}
\end{figure}

Our experiments are consistent with this intuition. In Figure \ref{fig:accuracy_beta}, we show that accuracy suffers \arx{when training with gradient penalties} as we increase $\beta$ but is unaffected for SGD. (We note that large $\beta$ does \emph{worse} than
ReLU due to the non-zero second derivative of $\beta$-GELU at $0$.)
% \arx{We also confirm experimentally that $\beta$ effectively controls the sparsity of the activation function Hessian in the appendix.}

% And in Figure \ref{fig:sparsity_beta} we confirm that $\beta$ effectively controls the sparsity of the activation function Hessian both at initialization and after training. This is evidence that the difference between the different activation functions for penalty \texttt{SAM} is explained by the statistics of the \mshort{} matrix.

% \begin{figure}[ht!]
%      \centering
%      \begin{subfigure}[b]{0.45\textwidth}
%          \centering
%          \includegraphics[width=\textwidth]{images/imagenet_sparsity_beta.pdf}
%          \caption{Imagenet}
%          \label{fig:y equals x}
%      \end{subfigure}
%      \hfill
%      \begin{subfigure}[b]{0.45\textwidth}
%          \centering
%          \includegraphics[width=\textwidth]{images/cifar10_sparsity_beta.pdf}
%          \caption{CIFAR-10}
%          \label{fig:three sin x}
%      \end{subfigure}
%         \caption{Fraction of nonzeroes in $\nabla^2_x\sigma(x)$ as $\beta$ increases across datasets for networks with $\beta$-GELU activations (average of 2 seeds). We can see that the sparsity of this second derivative increase dramatically as $\beta$ increases.}
%         \label{fig:sparsity_beta}
% \end{figure}

Note that we are not claiming that the choice of the activation function is a sufficient condition for gradient penalties to work with larger $\rad$. There are many architectural changes that can affect the \mshort{} matrix and we have shown that the statistics of the activation function is a significant one.

\arx{

\subsection{Augmented ReLU and diminished GELU}

\begin{figure}[ht!]
     \centering
     \begin{subfigure}[b]{0.45\textwidth}
         \centering
         \includegraphics[width=\textwidth]{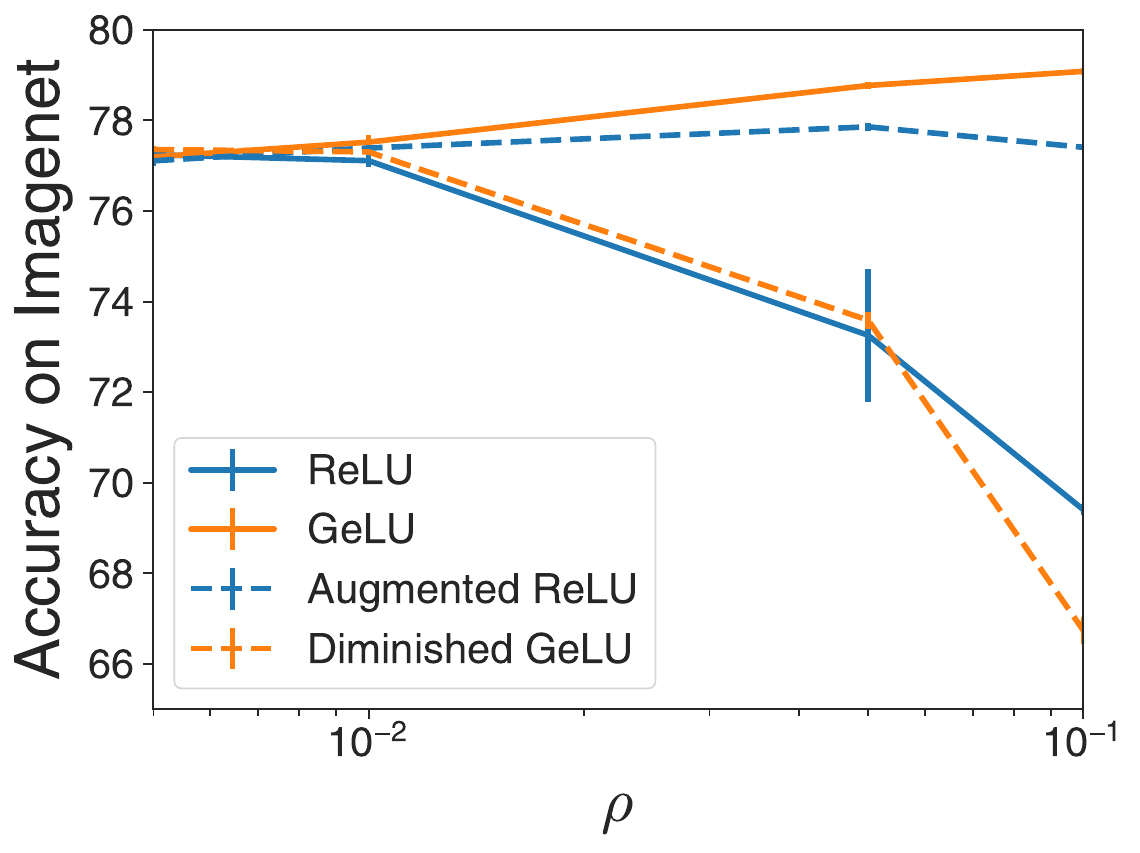}
         \caption{Imagenet}
         \label{fig:y equals x}
     \end{subfigure}
     \hfill
     \begin{subfigure}[b]{0.45\textwidth}
         \centering
         \includegraphics[width=\textwidth]{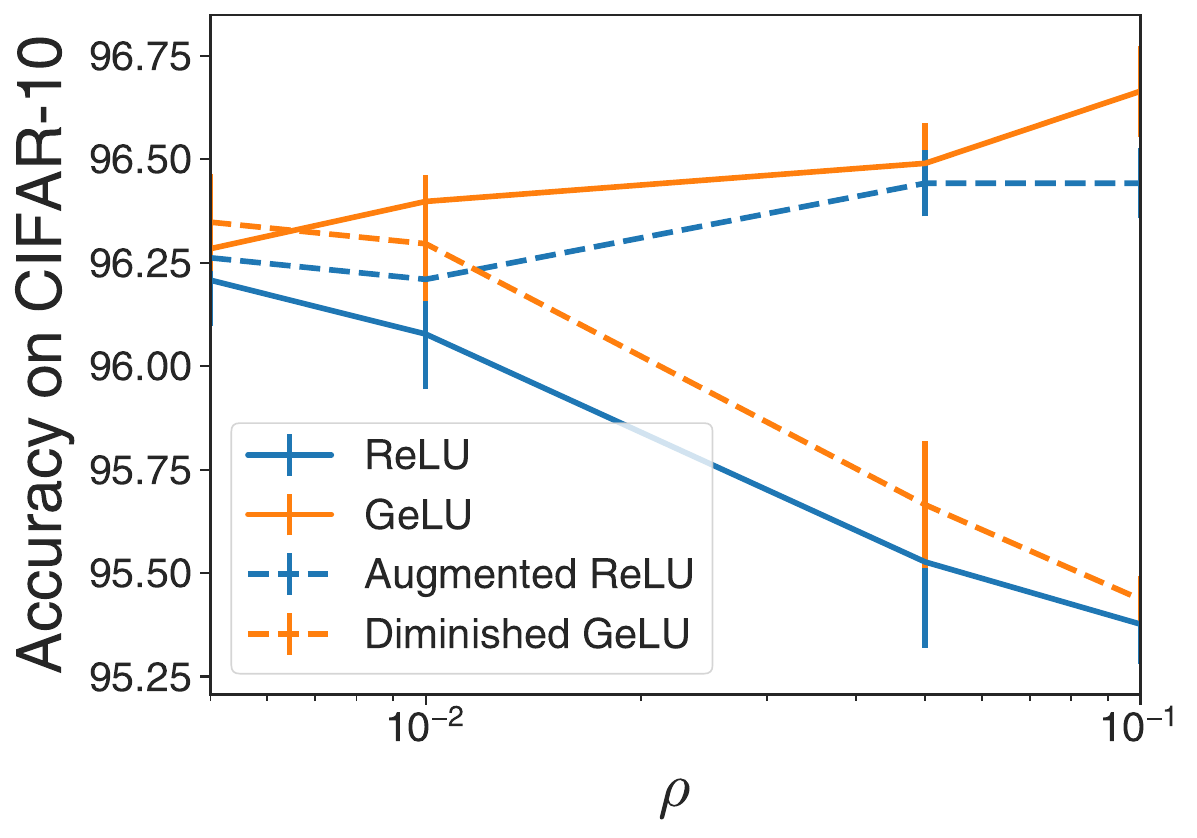}
         \caption{CIFAR-10}
         \label{fig:three sin x}
     \end{subfigure}
        \caption{Test accuracy as $\rho$ increases for Augmented ReLU and Diminished GELU (average of $5$ seeds). The addition or removal of information from the \mshort{} controls the effectiveness of the gradient penalty.}
        \label{fig:decoupled}
\end{figure}

We can perform a more direct experiment probing the effects of the second derivative part of
the \mshort{} on the learning dynamics by defining the \emph{augmented ReLU} and the
\emph{diminished GELU}. The basic idea is to design a modified ReLU which has
a well-posed second derivative, and to define a GELU that has a second derivative of $0$.
This lets us ``turn on'' the second derivative part of the \mshort{} for ReLU,
which previously had none, and ``turn off'' the second derivative part of
GELU, making it more similar to the setting with vanilla ReLU.

We will define our augmented and diminished functions using the ability to define custom
derivative functions in modern automatic differentiation (AD) frameworks. In AD frameworks,
the chain rule is decomposed into derivative operators
on basic functions combined with elementary operations. Let us denote the AD derivative
operator applied to function $f$ as $\dad[f]$. Normally this transformation corresponds to
the real derivative operator; that is, $\dad[f] := df/dx$.

However we can instead define a custom derivative $\dad[f](x) := g(x)$. The net result is that
any chain rule term evaluating $df/dx$ will be replaced by evaluation of $g$ at that
point. In this example, a second application of the AD operator nets us
$\dad[\dad[f]](x) = \dad[g](x)$ - which itself can be a custom derivative.
% This process is commonly used to define the derivatives of piecewise smooth
% functions like ReLU.

We define the augmented
ReLU as follows: $f_{aug}(x) := \relu(x)$ as normal. We make the common
choice to define the first AD
derivative as $\dad[f_{aug}](x) := \Theta(x)$,
the Heaviside step function ($\Theta(x) = 1$ if $x >0$, $\Theta(x) = 0$ otherwise).
The second AD derivative $\dad[\dad[f_{aug}]](x) = \dad[\Theta(x)]$.
Normally in AD frameworks, $\dad[\Theta(x)] := 0$ and therefore $\relu$ implementations
have no second derivative; we instead make the definition:
\begin{equation}
\dad[\dad[f_{aug}]](x) := \frac{\beta}{\sqrt{2\pi}}e^{-\beta^{2}x^{2}/2}
\end{equation}
Therefore the AD program replaces any second derivatives of $f_{aug}$ (e.g. during
HVP calculations) with a
Gaussian of width $\beta$, which approximates the delta
function in the limit $\beta\to\infty$. However for $\beta$ of $O(1)$, this gives
an approximation of the delta function that is more numerically stable, and lets
us test if gradient penalty with ReLU can be rescued by adding information related
to the second derivative piece of the \mshort{}.

Analogously, the diminished GELU is defined by ``turning off'' the second derivative of
GELU. Defining $f_{dim}(x):= \GELU(x)$, the first derivative is defined normally as
$\dad[f_{dim}](x) := g(x)$ where $g = d\GELU/dx$.
We define the AD derivative $\dad[g](x)$ to be $0$, which means:
\begin{equation}
\dad[\dad[f_{dim}]](x) := 0
\end{equation}
This brings the properties of GELU closer to that of ReLU at least in terms of the
higher order derivatives. Diminished GELU lets us test whether or not the second
derivative part of the \mshort{} is necessary for the success of gradient penalties
with GELU.

We used our Imagenet setup to train with augmented ReLU and diminished GELU (Figure \ref{fig:decoupled}).
We find that augmented ReLU performs better than plain ReLU and
nearly matches the performance of GELU, while diminished
GELU has poor performance similar to ReLU. This suggests that second derivative
information is necessary for the improved performance of GELU with gradient penalties,
and moreover it is helpful to make gradient penalties work with ReLU. This gives us
direct evidence that in this setting, information from the \mshort{} is crucial for
good generalization, and gradient penalties are sensitive to second derivatives
of activation functions.
}

\section{\arx{How \mshort{} affects training with Hessian penalties}}\label{sec:weight_noise}

% \aga{An alternative approach is to frame this as ``how \mshort{} affects training
% with Hessian penalties", and then say that we build the analysis around the classic
% example of weight noise. This gives us a better parallel with section 4, and emphasize
% the generality of our work - even if the reader doesn't care about weight noise,
% Hessian penalties are a natural progression from gradient penalties.
% }

\arx{
In this section we will show that the \mshort{} has significant impact on the effectiveness of Hessian penalties. In particular, we consider the case of weight noise because as we will see it is an efficient way to penalize the Hessian. In contrast to the previous section where the \mshort{} solely influenced learning dynamics, weight noise implicitly regularizes the \mshort{}. We will show through ablations that this regularization is detrimental and explain why.
}
% \aga{Is there a way to make a more positive statement as well? That via GN penalty 
% experiments we find a way to make the ``inspiration'' of weight noise work?}

% \reb{In the previous section we studied \emph{update rules} that involved Hessian-vector
% products; in this section we investigate \emph{regularizers} of the Hessian.}

% \reb{We center our investigation on a basic question:
% why does weight noise fail to improve generalization,
% }
% % We investigate why weight noise does not work
% even though it 
% has long been thought to be equivalent to a gradient penalty 
% similar to \ref{eqn:penalty_sam} \citep{bishop1995training}? 
% We will see that this connection does not hold for non-linear 
% models due to the existence of the \mshort{} term of the 
% Hessian.
% \reb{ Gradient descent on the \mshort{} term causes the update rule to involve 
% \emph{third derivatives} instead of Hessian-vector products alone. We will show that this leads to poor generalization
% and we are better off regularizing the GN term only (second derivatives in update equation, like penalty SAM).
% }

\subsection{\arx{Weight Noise analysis neglects the \mshort{}}}

We first review the analysis of training with noise established by \cite{bishop1995training}. Though the paper considers input noise, the same analysis can be applied to weight noise. Adding Gaussian $\ep \sim \mathcal{N}(0, \sigma^2)$ noise with strength hyper-parameter $\sigma$ to the parameters can be approximated to second order by 
\begin{equation}\label{eqn:wn_base}
    \expect_{\ep}[\Lo(\th + \ep)] \approx \Lo(\th) + \cancel{\expect_{\ep}[\nabla_{\th}\Lo\cdot\ep]} + \expect_{\ep}[\ep^{\tpose} \Hmat \ep] = \Lo(\th)+ \sigma^2\text{tr}(\Hmat)
\end{equation}
where the second term has zero expectation since $\ep$ is mean $0$,
and the third term is a variation of the Hutchison trace estimator \citep{hutchinson1989stochastic}.
(We note that though the second term vanishes in expectation, it still can
have large effects on the training dynamics.)
 \citep{bishop1995training} argues that we can simplify the term related to the Hessian by dropping the \mshort{} in Equation \ref{eqn:hess_decomp} for the purposes of minimization
\begin{equation}
    \text{tr}(\Hmat) = \text{tr}\left(\J^{\tpose} \Hmat_{\z} \J + \nabla_{\z}\Lo \cdot \nabla^{2}_{\th}\z\right) \approx \text{tr}(\J^{\tpose} \Hmat_{\z} \J)
\end{equation}
The argument is that for the purposes of training neural networks this term can be dropped because it is zero at the global minimum. 
% For models with mean squared error loss (MSE), this gives us
% \begin{equation}
%     E[\Lo_\text{MSE}(\th + \ep)] \approx \Lo_\text{MSE}(\th)+ \sigma^2\text{tr}(\J^{\tpose} \Hmat^\text{MSE}_{\z} \J) = \Lo_\text{MSE}(\th)+ \sigma^2\left\|\nabla_{\th} \z\right\|^2
% \end{equation}
% This is strikingly similar to the penalty form of \texttt{SAM} (Equation 
% \ref{eqn:penalty_sam}), with the key difference being that it is
% a gradient penalty on the logits and not the loss. In fact for MSE there is \emph{no} information about the loss in this term.
% Recent work has proposed a new estimator for the trace of the Gauss-Newton
% matrix for cross-entropy loss \cite{wei2020implicit}.
% Using this 
% estimator, we can express weight noise 
% with cross-entropy loss as \aga{We should make clear that this is the "target" of weight 
% noise or something like that - here there is no NME.}
% \begin{equation}\label{eqn:wn_ce}
%     \Lo_\text{CE}(\th + \ep) \approx \Lo_\text{CE}(\th) + \sigma^2\text{tr}(\J^{\tpose} \Hmat^\text{CE}_{\z} \J) =  \Lo_\text{CE}(\th) + \sigma^2 \expect_{\hat{\y}\sim \text{Cat}(\z)}\left[\left\|\nabla_{\th}\Lo (\th, \hat{\y})\right\|^2\right].
% \end{equation}
% This is almost exactly the same as Equation \ref{eqn:gradient_penalty}, except for 
% the fact that the labels are sampled from the model instead of the ground-truth labels.

\arx{However, the hypothesis that the \mshort{} has negligible impact in this setting has not been experimentally verified. We address this gap in the next section by providing evidence that the \mshort{} cannot be neglected for modern networks.
}

% In Section \ref{sec:wn_variants} we
% design a series of experiments to probe these questions, and provide evidence that
% the \mshort{} cannot be neglected for modern networks, and there is a difference
% between penalizing the gradients of $\Lo$ and penalizing the gradients of
% $\z$.

% Despite the close similarity between Equation \ref{eqn:penalty_sam_unnorm} and 
% \ref{eqn:wn_ce}, we will see in Section \ref{sec:wn_variants} that they have significantly different effects on generalization. This leads us to question if the \mshort{}
% is in fact negligible in deep learning.
% Indeed, the contribution of the \mshort{} term may have 
% been negligible for networks that do not perform much feature learning, but that 
% is not the case for modern networks.

\subsection{\arx{Ablations reveal the influence of the \mshort{}}}\label{sec:wn_variants}

\begin{figure}[ht]
     \centering
     \begin{subfigure}[b]{0.45\textwidth}
         \centering
         \includegraphics[width=\textwidth]{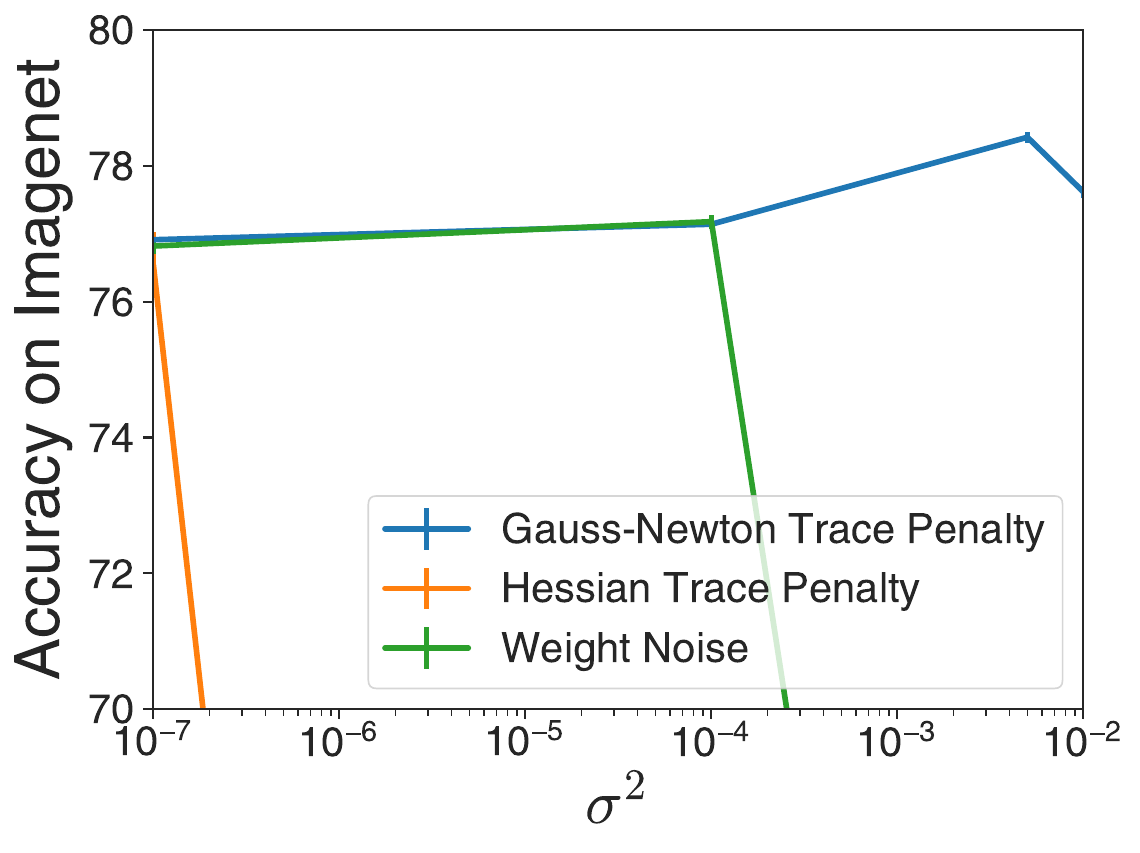}
         \caption{Imagenet}
         \label{fig:y equals x}
     \end{subfigure}
     \hfill
     \begin{subfigure}[b]{0.45\textwidth}
         \centering
         \includegraphics[width=\textwidth]{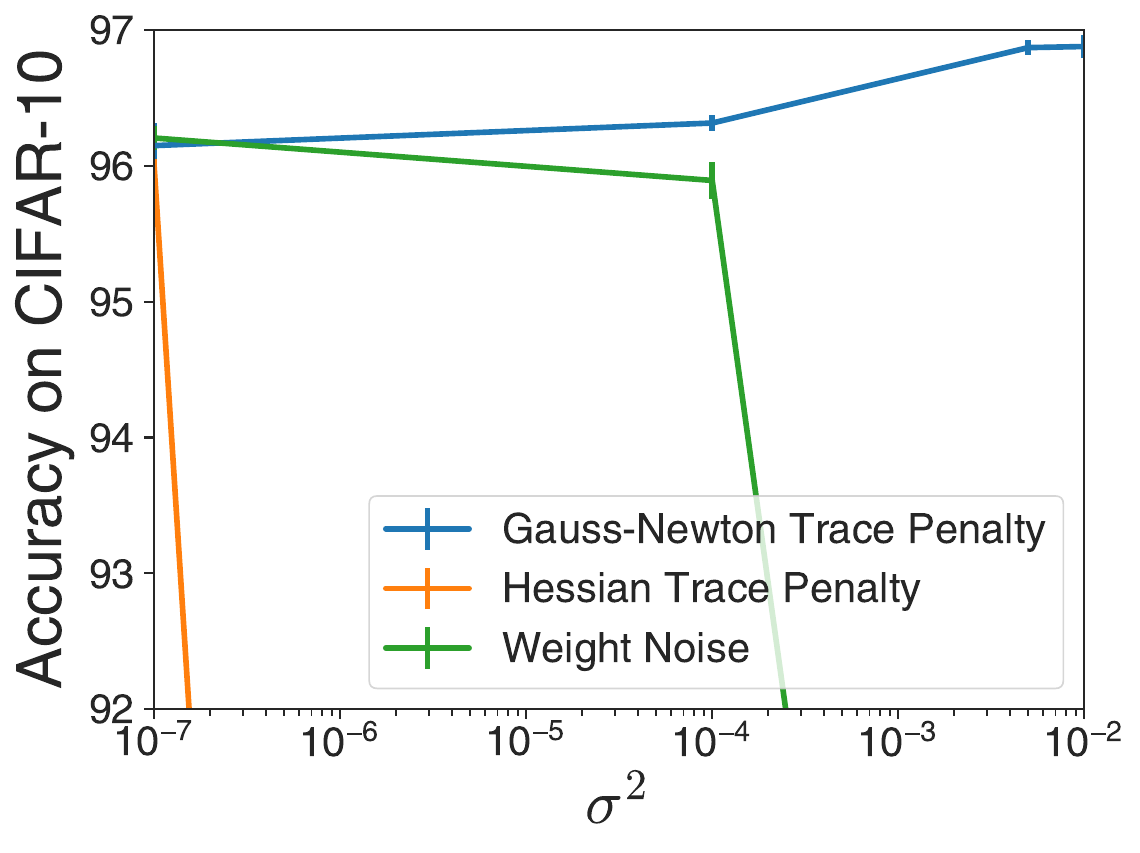}
         \caption{CIFAR-10}
         \label{fig:three sin x}
     \end{subfigure}
        \caption{Test Accuracy as $\sigma^2$ increases across different datasets and activation functions averaged over $5$ seeds. Large $\sigma^2$ reveals a stark contrast between the Gauss-Newton trace penalty (excluding \mshort{}) and methods incorporating it, highlighting the \mshort{}'s influence.}
        \label{fig:mystery2_rho}
\end{figure}

\arx{In order to study the impact of the \mshort{} in this setting, we evaluate ablations of weight noise to determine the impact of the different components. Recalling Equation \ref{eqn:wn_base}, the methods we will consider are given by
\begin{equation}
        \underbrace{\expect_{\ep}[\Lo(\th + \ep)]}_{\text{Weight Noise}}  = \overbrace{\underbrace{\Lo(\th) + \sigma^2\text{tr}\left(\J^{\tpose} \Hmat_{\z} \J\right)}_{\text{Gauss-Newton Trace Penalty}} + \sigma^2\text{tr}\left(\nabla_{\z}\Lo \cdot \nabla^{2}_{\th}\z\right)}^{\text{Hessian Trace Penalty}} + \mathcal{O}(\|\ep\|^2)
\end{equation}

{\bf Hessian Trace penalty}\quad This ablation allows 
    to us to single out the second order effect of weight noise, as it's possible
    the higher order terms from weight noise affect generalization. We implement this penalty with Hutchinson's trace estimator ($\text{tr}(\Hmat)= \expect_{\ep\sim\mathcal{N}(0, 1)}[\ep^T \Hmat \ep]$).
    
{\bf Gauss-Newton Trace penalty}\quad This ablation removes the \mshort{}'s contribution, enabling us to isolate and measure its specific influence on the model.
Recent work has proposed a new estimator for the trace of the Gauss-Newton
matrix for cross-entropy loss \cite{wei2020implicit}.
Using this 
estimator, we can efficiently compute this penalty using
\begin{equation}
    \text{tr}\left(\J^{\tpose} \Hmat_{\z} \J\right)=\expect_{\hat{\y}\sim \text{Cat}(\z)}[\left\|\nabla_{\th}\Lo (\th, \hat{\y})\right\|^2]
    \label{eq:gn_estimator}
\end{equation}
where $\text{Cat}(\cdot)$ is the categorical distribution and $\z$ are the logits. This computes the norm of the gradients, but with the labels sampled from the model instead of the ground-truth.  We 
    do not pass gradients through the sampling of the labels $\hat{\y}$, but we find similar 
    results if we pass gradients using the straight-through estimator 
    \citep{bengio2013estimating}. Note the similarity to the gradient penalties studied in the previous section, which we will address in later sections.
% This is almost exactly the same as Equation \ref{eqn:gradient_penalty}, except for 
% the fact that the labels are sampled from the model instead of the ground-truth labels.

We  draw a single sample to estimate the expectations for the different estimators. We experimented with 2 samples for the Hessian Trace penalty but we found this did not affect the results. 

Figure \ref{fig:mystery2_rho} shows that the methods perform quite differently as $\sigma^2$ increases - confirming the influence of the \mshort{}. We can see that the generalization improvement of the Gauss-Newton Trace penalty
is consistently greater than either weight noise or Hessian Trace penalty. Its improvement on Imagenet is a 
significant $1.6\%$. In contrast, the other methods provide little accuracy improvement. 
}

These results are evidence that the \mshort{} term of the Hessian should not be 
dropped when applying the \emph{analysis} of \citep{bishop1995training} to weight noise for
modern networks. Indeed there 
is a significant difference between the Hessian trace penalty, which involves
the \mshort{}, and the Gauss-Newton penalty, which 
does not. \arx{This suggests that while the \mshort{} has a positive influence on the learning dynamics as seen for gradient penalties in Section \ref{sec:grad_pen}, it is detrimental to regularize it directly in the loss function.
}

\arx{This is not contradictory with the analysis in Section \ref{sec:grad_pen}
which suggested that incorporating \mshort{} information into updates helps
learning. Minimizing the \mshort{} through the loss will reduce
its impact on the learning dynamics. We can also see that the Gauss-Newton penalty,
which does not involve the \mshort{} in the loss, indeed involves the \mshort{}
in the update rule:
\begin{equation}
\nabla_{\th}\tr\left(\J^{\tpose} \Hmat_{\z} \J\right) = \expect_{\hat{\y}\sim \text{Cat}(\z)}[\left(\nabla_{\z}\Lo(\th,\hat{\y})\cdot\nabla_{\th}^{2}\z\right)\nabla_{\th}\Lo(\th, \hat{\y})]
\end{equation}
This update rule is very similar to the update rule for the gradient penalty
in Equation \ref{eq:gpen_update}. The three differences are the lack of a normalization
factor (equivalent to $p = 2$ gradient penalty), the lack of Gauss-Newton vector product,
and the fact that the \mshort{} is computed over the labels generated by the model
and not the true labels. Therefore the Gauss-Newton trace penalty, the best performing
of our ablations, does indeed incorporate \mshort{} information into the update rule.
}

% \aga{Can probably cut next equation; not sure it adds much}

% \arx{

% In contrast, the update rule for an NME trace penalty looks like
% \begin{equation}
% -\lr\nabla_{\th}\left[\rad\tr\left(\nabla_{\z}\Lo \cdot \nabla^{2}_{\th}\z\right)\right] = -\lr\rad \left(\J^{\tpose}\Hmat_{\z}\tr\left[\nabla_{\th}^{2}\z\right]+\nabla_{\z}\Lo\cdot\nabla_{\th}\tr[\nabla^{2}_{\th}\z]\right)
% \end{equation}
% which is qualitatively different from any gradient penalty.

% }

% We can see in Figure \ref{fig:sparsity_variants} that the Hessian is indeed getting penalized between Gauss-Newton and Weight noise from the evolution of the activation Hessian during training (try to match the coefficient)

% \begin{figure}[ht!]
%      \centering
%      \begin{subfigure}[b]{0.45\textwidth}
%          \centering
%          \includegraphics[width=\textwidth]{images/imagenet_variants_sparsity_rho.pdf}
%          \caption{Imagenet}
%          \label{fig:y equals x}
%      \end{subfigure}
%      \hfill
%      \begin{subfigure}[b]{0.45\textwidth}
%          \centering
%          \includegraphics[width=\textwidth]{images/cifar10_variants_sparsity_rho.pdf}
%          \caption{CIFAR-10}
%          \label{fig:three sin x}
%      \end{subfigure}
%         \caption{Fraction of nonzeroes in $\nabla^2_x\sigma(x)$ as $\beta$ as $\rad$ increases for weight noise and the Gauss-Newton penalty across two datasets. Weight noise increases sparsity on Imagenet, but does not on CIFAR-10.}
%         \label{fig:sparsity_variants}
% \end{figure}

\arx{\section{Discussion}

Our theoretical analysis gives some understanding of the structure of the Hessian -
in particular, the \mname{} matrix. This piece of the Hessian is often neglected as it
is generally indefinite and doesn't generate large eigenvalues,
and is $0$ at an interpolating minimum. However, the \mshort{} can encode important
information related to feature learning
as it depends on $\nabla^{2}_{\th}\z$ - the gradient of the Jacobian.
For example, in networks with saturating activation functions the \mshort{}
gives information about the potential benefits of switching into the saturated
regimes of different neurons. More generally, our analysis shows
that the elements of the \mshort{}, especially on the diagonal
are sensitive to the second derivative of the activation function.

\subsection{\mshort{} and gradient penalties}

Our experiments suggest that these second derivative properties can be quite important
when training with gradient penalty regularizers. ReLU has a poorly defined pointwise
second derivative and the regularizer harms training,
while GELU has a well defined one and gains benefits from modest values of the regularizer.
Our experiments with $\beta$-GELU suggest that if the \mshort{} is well-defined but sparse
and ``spiky'', we also achieve poor training.

One important point here is that the sensitivity to the second derivative comes from
the fact that the update rule (Equation \ref{eq:gpen_update}) involves an explicit
Hessian-vector product. We can contrast this with methods which use second order
information \emph{implicitly} via first order measurements. In particular, the
\texttt{SAM} algorithm for controlling curvature is equivalent, to low order in
regularization strength $\rad$, to gradient penalties (Appendix \ref{app:sam_and_gpen}).
\texttt{SAM} approximates dynamics on a sharpness-penalized objective by
taking two steps with gradient information, and it
works with both ReLU and GELU - matching
the performance of gradient penalties on GELU (Appendix \ref{app:psam_vs_osam}).

The difference between \texttt{SAM} and the gradient penalty is that \texttt{SAM}
acquires second order information via discrete, gradient-based steps. It is effectively
\emph{integrating} over the Hessian (and therefore \mshort{}) information. Therefore
it is not as sensitive to the pointwise properties of the second derivative of the
activation function.

The \mshort{} is also important in understanding the regularizers in Section
\ref{sec:weight_noise}, where we showed that even in the case of the Gauss-Newton
trace penalty, the \mshort{} shows up in the update rule. Therefore the \mshort{}
can be important for understanding dynamics even when regularization efforts
focus on the Gauss-Newton.

\subsection{Lessons for using second order information}

Our work suggests that some second order methods may benefit from tuning the \mshort{}.
This is especially true for methods which result in Hessian-vector products in update rules
(like the gradient and Hessian penalties studied here).
% This may be less relevant for second-order optimizers which focus on pre-conditioning of 
% directions associated with large eigenvalues (which are usually dominated by the
% GN rather than the \mshort{}).

Our experiments with augmented ReLU suggest that helpful interventions can be
designed to improve propagation of \mshort{} information.
We hypothesize that this information is related to feature learning,
and therefore acts over the totality of training to affect generalization.

}

\section{Conclusion}

Our work sheds light on the complexities of using second order information in deep
learning. \arx{We have identified clear cases where} it is important to consider the effects of \emph{both} the Gauss-Newton and
\mname{} terms, and design algorithms and architectures with that in mind.
Designing activation functions for compatibility with second order methods may also be
an interesting avenue of future research.

\clearpage

\bibliography{iclr2024_conference, atish_refs_iclr2024}
\bibliographystyle{iclr2024_conference}

\appendix

\section{Hessian structure}

\subsection{Gauss-Newton and NTK learning}

\label{app:ntk_gn}

In the large width limit (width/channels/patches increasing while dataset is fixed),
the learning dynamics of neural networks are well described by the
\emph{neural tangent kernel}, or NTK \citep{jacot_neural_2018, lee_wide_2019}. Consider a dataset size
$\D$, with outputs $\z(\th, \X)$ over the inputs $\X$ with parameters $\th$.
The (empirical)
NTK $\sm{\ntk}$ is the $\D\times\D$ matrix given by
\begin{equation}
\sm{\ntk} \equiv \frac{1}{\D}\J\J^{\tpose},~\J \equiv \frac{\partial \z}{\partial\th}
\end{equation}
For wide enough networks, the learning dynamics can be written in terms of
the model output $\z$ and the NTK $\ntk$ alone. For small learning rates
we can study the gradient flow dynamics. The gradient flow dynamics on the parameters
$\th$ with loss function $\Lo$ (averaged over the dataset) is given by
\begin{equation}
\dot{\th} = -\frac{1}{\D}\nabla_{\th}\Lo = -\frac{1}{\D}\J^{\tpose}\nabla_{\z}\Lo
\end{equation}
We can use the chain rule to write down the dynamics of $\z$:
\begin{equation}
\dot{\z} = \frac{\partial\z}{\partial\th}\dot{\th} = -\frac{1}{\D}\J\J^{\tpose}\nabla_{z}\Lo = -\sm{\ntk}\nabla_{z}\Lo
\label{eq:ntk_ode}
\end{equation}

In the limit of infinite width, the overall changes in individual parameters become
small, and the $\ntk$ is fixed during training. This corresponds to the \emph{linearized}
or \emph{lazy} regime \cite{chizat_lazy_2019, agarwala_temperature_2020}. The NTK encodes the
linear response of $\z$ to small changes in $\th$, and the dynamics is closed
in terms of $\z$.
For finite width networks, this can well-approximate the dynamics for a number
of steps related to the network width amongst other properties  \cite{lee_wide_2019}.

In order to understand the dynamics of Equation \ref{eq:ntk_ode} at small times,
or around minima, we can linearize with respect to $\z$. We have:
\begin{equation}
\frac{\partial\dot{\z}}{\partial \z} = -\frac{\partial \sm{\ntk}}{\partial\z}\nabla_{\z}\Lo-\sm{\ntk}\Hmat_{\z}
\end{equation}
where $\Hmat_{\z} = \frac{\partial^2\Lo}{\partial\z\partial\z'}$.
In the limit of large width, the NTK is constant and the first term vanishes.
The local dynamics depends on the spectrum of $\sm{\ntk}\Hmat_{\z}$. From the
cyclic property of the trace, the non-zero part of the spectrum is equal to the
non-zero spectrum of $\frac{1}{\D}\J^{\tpose}\Hmat_{\z}\J$ - which is the
Gauss-Newton matrix.

Therefore the eigenvalues of the Gauss-Newton matrix control
the short term, linearized dynamics of $\z$, for fixed NTK. It is in this sense
that the Gauss-Newton encodes information about exploiting the local linear
structure of the model.

\subsection{\mname{} and second derivatives of FCNs}

\label{app:second_deriv}

We can explicitly compute the Jacobian and second derivative of the model for a fully connected
network. We write a feedforward network as follows:
\begin{equation}
\h_{\l} = \W_{\l}\x_{\l},~\x_{\l+1} = \phi(\h_{\l})
\end{equation}
The gradient of $\x_{L}$ with respect to $\W_{l}$ can be written as:
\begin{equation}
\frac{\partial \x_{L}}{\partial \W_{\l}} = \frac{\partial \x_{L}}{\partial \h_{\l}}\frac{\partial \h_{l}}{\partial \W_{l}}
\end{equation}
which can be written in coordinate-free notation as
\begin{equation}
\frac{\partial \x_{L}}{\partial \W_{\l}} = \frac{\partial \x_{L}}{\partial \h_{\l}}\otimes\x_{l}
\end{equation}
If we define the partial Jacobian $\J_{\l'\l}\equiv \frac{\partial\x_{l'}}{\partial\x_{l}}$,
$\l'>\l$
\begin{equation}
\frac{\partial \x_{L}}{\partial \W_{\l}} = \J_{L(\l+1)}\circ\phi'(\h_{l}) \otimes\x_{l}
\end{equation}
Here $\circ$ denotes the Hadamard product, in this case equivalent to matrix multiplication by
$\diag(\phi'(\h_{m}))$. 

The Jacobian can be explicitly written as
\begin{equation}
\J_{\l'\l} = \prod_{m=\l}^{\l'-1}\phi'(\h_{m})\circ\W_{m}
\end{equation}
Therefore, we can write:
\begin{equation}
\frac{\partial \x_{L}}{\partial \W_{\l}} = \left[\prod_{m=\l+1}^{\L-1}\phi'(\h_{m})\circ\W_{m}\right]\circ\phi'(\h_{l})\otimes\x_{l}
\end{equation}

The second derivative is more complicated. Consider
\begin{equation}
\frac{\partial^2\x_{L}}{\partial \W_{\l}\partial\W_{m}} = \frac{\partial }{\partial \W_{m}}\left[
\J_{L(\l+1)}\circ \phi'(\h_{\l})\otimes \x_{l}\right]
\end{equation}
for weight matrices $\W_{\l}$ and $\W_{m}$. Without loss of generality,
assume $m\geq l$.

We first consider the case where $m>l$. In this case, we have
\begin{equation}
\frac{\partial \phi'(\h_{\l})}{\partial \W_{m}} = 0,~ \frac{\partial\x_{l} }{\partial \W_{m}} = 0
\end{equation}
since $\W_{m}$ comes after $\h_{l}$. If we write down the derivative of
$\J_{L(l+1)}$, there are two types of terms. The first comes from the direct
differentiation of $\W_{m}$; the others come from differentation of
$\phi'(\h_{n})$ for $n\geq m$. We have:
\begin{equation}
\frac{\partial \J_{L(l+1)}}{\partial \W_{m}} = \J_{L(m+1)}\phi'(\h_{m})\frac{\partial\W_{m}}{\partial \W_{m}}\J_{(m-1)(l+1)}+\sum_{o = m}^{L-1}\J_{L(o+1)}
\frac{\partial \phi'(\h_{o})}{\partial \W_{m}}\W_{o}\J_{(o-1)(l+1)}
\end{equation}

The $\W_{m}$ derivative projected into a direction $\Bm$ can be written as:
\begin{equation}
\begin{split}
\frac{\partial \J_{L(l+1)}}{\partial \W_{m}}\cdot\Bm & = \J_{L(m+1)}\phi'(\h_{m})\Bm\J_{(m-1)(l+1)}\\
& +\sum_{o = m}^{L-1}\J_{L(o+1)}\left(\phi''(\h_{o})\circ \W_{o}\frac{\partial \x_{o-1}}{\partial\W_{m}}\cdot\Bm \right)
\W_{o}\J_{(o-1)(l+1)}
\end{split}
\end{equation}
From our previous analysis, we have:
\begin{equation}
\begin{split}
\frac{\partial \J_{L(l+1)}}{\partial \W_{m}}\cdot\Bm & = \J_{L(m+1)}\phi'(\h_{m})\Bm\J_{(m-1)(l+1)}\\
& +\sum_{o = m}^{L-1}\J_{L(o+1)}\left(\phi''(\h_{o})\circ \left[\W_{o}\J_{o(m+1)}\circ\phi'(\h_{m+1})\circ\Bm\x_{m}\right]\right)
\frac{\partial \phi'(\h_{o})}{\partial \W_{m}}\W_{o}\J_{(o-1)(l+1)}
\end{split}
\end{equation}
In total, the second derivative projected into the $(\Am, \Bm)$ direction for
$m > \l$ is given by:
\begin{equation}
\begin{split}
\frac{\partial^2 \x_{L}}{\partial\W_{l}\partial \W_{m}}\cdot(\Am\otimes\Bm) & = \left[\J_{L(m+1)}\phi'(\h_{m})\Bm\J_{(m-1)(l+1)}+\right.\\
& \left.\sum_{o = m}^{L-1}\J_{L(o+1)}\left(\phi''(\h_{o})\circ \left[\W_{o}\J_{o(m+1)}\circ\phi'(\h_{m+1})\circ\Bm\x_{m}\right]\right)
\frac{\partial \phi'(\h_{o})}{\partial \W_{m}}\W_{o}\J_{(o-1)(l+1)}\right]\\
& \circ \phi'(\h_{\l})\Am \x_{l}
\end{split}
\end{equation}

Now consider the case $m = \l$. Here there is no direct differentiation with
respect to $\W_{m}$, but there is a derivative with respect to $\phi'(\h_{m})$.
The derivative is written as:
\begin{equation}
\begin{split}
\frac{\partial^2 \x_{L}}{\partial\W_{m}\partial \W_{m}}\cdot(\Am\otimes\Bm) & =
\J_{L(m+1)}\circ [\phi''(\h_{m})\circ \Bm\x_{l}] \Am\x_{m}+
\\
& \left[\sum_{o = m}^{L-1}\J_{L(o+1)}\left(\phi''(\h_{o})\circ \left[\W_{o}\J_{o(m+1)}\circ\phi'(\h_{m+1})\circ\Bm\x_{m}\right]\right)
\frac{\partial \phi'(\h_{o})}{\partial \W_{m}}\W_{o}\J_{(o-1)(m+1)}\right]\\
& \circ \phi'(\h_{m})\Am \x_{m}
\end{split}
\end{equation}

There are two key points: first, all but one of the terms in the off-diagonal
second derivative depend on only first derivatives of the activation; for
a deep network, the majority of the terms depend on $\phi''$. Secondly,
on the diagonal, all terms depend on $\phi''$. Therefore if $\phi''(x) = 0$,
the diagonal of the model second derivative is $0$ as well.

\section{\texttt{SAM} and gradient penalties}

\label{app:sam_and_gpen}

The gradient penalties studied in Section \ref{sec:grad_pen} are related to the Sharpness
Aware Minimization algorithm (\texttt{SAM}) developed to combat high curvature in deep
learning \citep{foret2020sharpness}. In this appendix we review the basics of \texttt{SAM},
show the correspondence to gradient penalties, and show that \texttt{SAM} is less
sensitive to the choice of activation function.

\subsection{\texttt{SAM}}

The ideas behind the \texttt{SAM} algorithm originates from seeking a minimum with a \emph{uniformly low loss} in its neighborhood (hence flat). This is formulated in \cite{foret2020sharpness} as a minmax problem,
\begin{equation}
\min_{\th} \max_{\boldsymbol{\epsilon}} \Lo (\th + \boldsymbol{\epsilon}) \quad\mbox{s.t.}\quad \|\boldsymbol{\epsilon}\| \leq \rad \,.
\end{equation}
For computational tractability, \cite{foret2020sharpness} approximates the inner
optimization by linearizing $\Lo$ w.r.t. $\boldsymbol{\epsilon}$ around the origin. Plugging the optimal
$\sm{\epsilon}$ into the objective function yields
\begin{equation}
\label{eq:SAM_Ideal}
\min_{\th} \Lo\Big(\th + \rad \, \frac{\nabla_{\th} \Lo(\th)}{\|\nabla_{\th} \Lo(\th)\|}\Big) \,.
\end{equation}
To minimize the above by gradient descent, we would need to compute\footnote{In our notation the gradient and Hessian operators $\nabla$ and $\nabla^2$ precede function evaluation, e.g. $\nabla_{\th}\Lo(f(\th))$ means $\big(\frac{\partial}{\partial \boldsymbol{\tau}}\Lo(\boldsymbol{\tau})\big)_{\boldsymbol{\tau}=f(\th)}$.}:
\begin{equation}
\label{eq:grad_of_SAM}
\frac{\partial}{\partial \boldsymbol{\theta}} \Lo\Big(\th + \rad  \frac{\gvec(\th)}{\|\gvec(\th)\|}\Big) \,=\,  \Bigg( \m{I} + \underbrace{\rad \frac{\Hmat}{\| \gvec\|} \Big( \m{I} - \frac{\gvec}{\|\gvec\|} \frac{{\gvec}^{\tpose}}{\|\gvec\|}\Big)}_{\mbox{Hessian related term}} \Bigg) \,\,\, \nabla_{\th} \Lo\left(\th+\rad \frac{\gvec}{\|\gvec\|}\right) 
 \,,~\gvec\equiv \nabla_{\th}\Lo(\th),~\Hmat \equiv \nabla_{\th}^2\Lo(\th)
\end{equation}
This can still be computationally
demanding as it involves the computation of a Hessian-vector product $\Hmat \gvec$. The
\texttt{SAM} algorithm drops the Hessian related term in (\ref{eq:grad_of_SAM}) giving
the update rule:
% \begin{equation}
% \th \leftarrow \th - \eta \, \left.\nabla_{\th} \Lo\right|_{\th+\rad\tilde{\gvec}},~\tilde{\gvec}\equiv \left.\nabla_{\th}\Lo\right|_{\th}/||\left.\nabla_{\th}\Lo\right|_{\th} ||
% \label{eq:sam_update_rule}
% \end{equation}
\begin{equation}
\th \leftarrow \th - \eta \, \nabla_{\th} \Lo\left(\th+\rad\tilde{\gvec}\right),~\tilde{\gvec}\equiv \nabla_{\th}\Lo(\th)/||\nabla_{\th}\Lo(\th) ||
\label{eq:sam_update_rule}
\end{equation}
for some step-size parameter $\eta>0$.  A related learning algorithm
is unnormalized \texttt{SAM} (\texttt{U\texttt{SAM}}) with update rule \citep{andriushchenko2022towards}
\begin{equation}
\th \leftarrow \th - \eta \, \nabla_{\th} \Lo\left(\th+\rad\gvec\right),~\gvec\equiv \nabla_{\th}\Lo(\th)
\label{eq:usam_update_rule}
\end{equation}
\texttt{U\texttt{SAM}} has similar performance to \texttt{SAM} and is easier to analyze
\citep{agarwala_sam_2023}.

% \begin{equation}
% \th \leftarrow \th - \eta \, \Bigg( \nabla_{\th}\Lo\Big(\th + \rad \, \frac{\nabla_{\th}\Lo(\boldsymbol{u})}{\|\nabla_{\th}\Lo(\boldsymbol{u})\|}\Big) \Bigg)_{\boldsymbol{u}=\th}\,,
% \end{equation}
% % \begin{equation}
% % \th \leftarrow \th - \eta \, \Bigg( \nabla_{\th}\Lo\Big(\boldsymbol{u}\Big) \Bigg)_{\th + \rad \, \frac{\nabla_{\th}\Lo(\boldsymbol{\th})}{\|\nabla_{\th}\Lo(\boldsymbol{\th})\|}}\,,
% % \end{equation}

% \begin{equation}
% \th \leftarrow \th - \eta \, \left.\nabla_{\th}\Lo \right|_{\theta+\rad \gvec},~\gvec = \left.\nabla_{\th}\Lo \right|_{\theta}
% \end{equation}

% which can be presumed as minimizing the following objective function,
% \begin{equation}
% \label{eq:sam_objective}
% \Lo_\text{\texttt{SAM}}(\th) \,\triangleq\, \Lo\Big(\th + \rad \, \texttt{stop\_grad}(\frac{\nabla_{\th}\Lo(\th)}{\|\nabla_{\th}\Lo(\th)\|})\Big) \,.
% \end{equation}

\subsection{Penalty \texttt{SAM}}

If $\rad$ is very small, we may approximate $\Lo$ in (\ref{eq:SAM_Ideal}) by its first order Taylor expansion around the point $\rad=0$ as below.
\begin{align}
\Lo_\text{PSAM}(\th) \,&\triangleq\, \Lo(\th)_{\rad=0} + \rad \Big(\frac{\partial}{\partial \rad} \Lo \, \Big(\th + \rad  \frac{\nabla_{\th}\Lo(\th)}{\|\nabla_{\th}\Lo(\th)\|}\Big) \Big)_{\rad=0} \,+O(\rad^{2})\\
& =\, \Lo(\th) + \rad \left\langle \nabla_{\th}\Lo(\th)  \,,\, \frac{\nabla_{\th}\Lo(\th)}{\|\nabla_{\th}\Lo(\th)\|} \right\rangle+O(\rad^{2}) \\
& \,=\, \Lo(\th) + \rad \, \|\nabla_{\th}\Lo(\th)\| +O(\rad^{2})\,.\label{eqn:penalty_sam}
\end{align}
Dropping terms of $O(\rad^{2})$ we arrive at the gradient penalty with $p = 1$.
If $\rad$ is not close to zero, 
then loss landscape of $\Lo_\text{PSAM}$ provides a very poor approximation to that of 
\ref{eq:SAM_Ideal}. In the remainder of this section,
we refer to this specific gradient penalty as \emph{Penalty SAM} and 
denote its associated objective function (\ref{eqn:penalty_sam}) by \texttt{P\texttt{SAM}}. The unnormalized equivalent \texttt{PU\texttt{SAM}} is
\begin{equation}
\Lo_\text{PUSAM}(\th) \,\triangleq\, \Lo(\th) + \rad \, \|\nabla_{\th}\Lo(\th)\|^2 +O(\rad^{2})\,.
\label{eqn:penalty_sam_unnorm}
\end{equation}
which corresponds to the $p = 2$ case of the gradient penalty.
% \texttt{PU\texttt{SAM}} was also studied in \cite{smith2021origin}, where
% analysis showed that gradient flow with penalty \texttt{SAM} and $\rad = \lr/4$ was
% a better approximator of \texttt{SGD} dynamics than pure gradient flow. In general the
% effects \texttt{PU\texttt{SAM}} can't be explained with an equivalent larger learning
% rate; see Appendix \ref{app:pen_vs_step_size} for more details.

\subsection{Penalty \texttt{SAM} vs Original \texttt{SAM}}

\label{app:psam_vs_osam}

\begin{figure}[ht]
     \centering
     \begin{subfigure}[b]{0.45\textwidth}
         \centering
         \includegraphics[width=\textwidth]{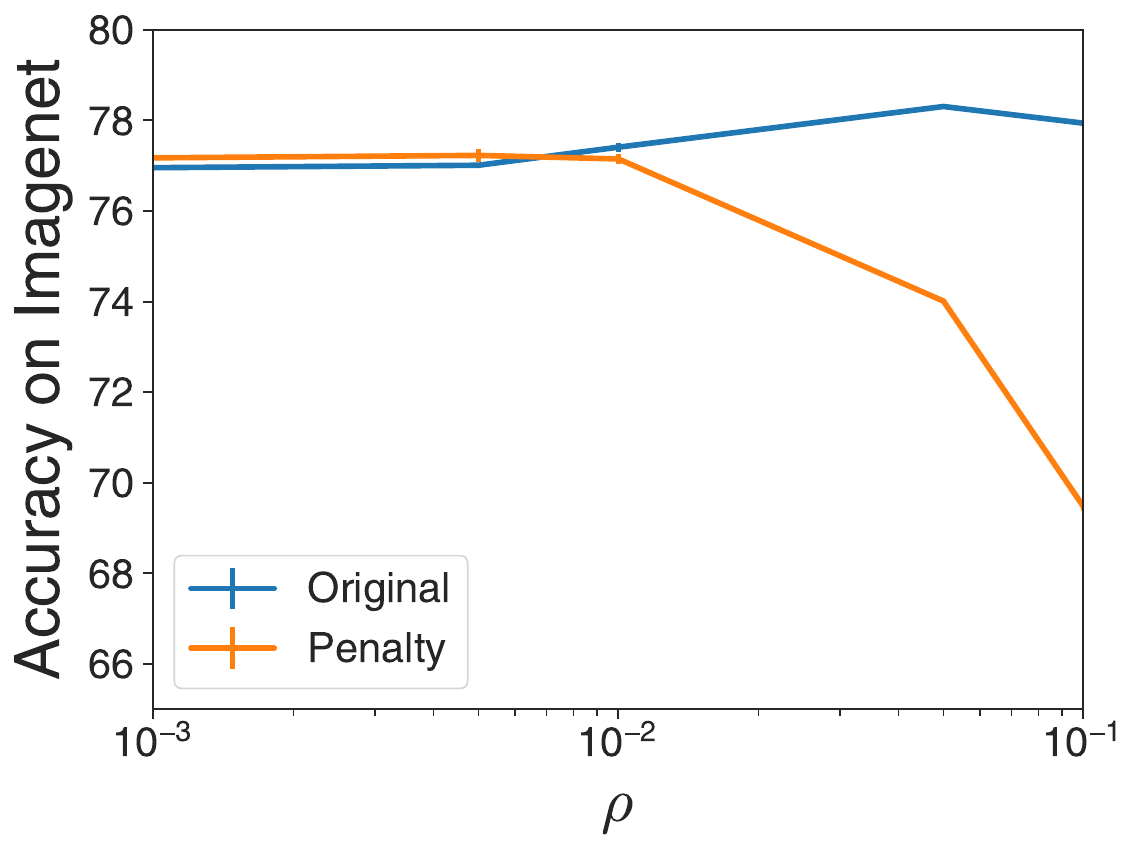}
         \caption{Imagenet with ReLU}
         \label{fig:y equals x}
     \end{subfigure}
    \hfill
     \begin{subfigure}[b]{0.45\textwidth}
         \centering
         \includegraphics[width=\textwidth]{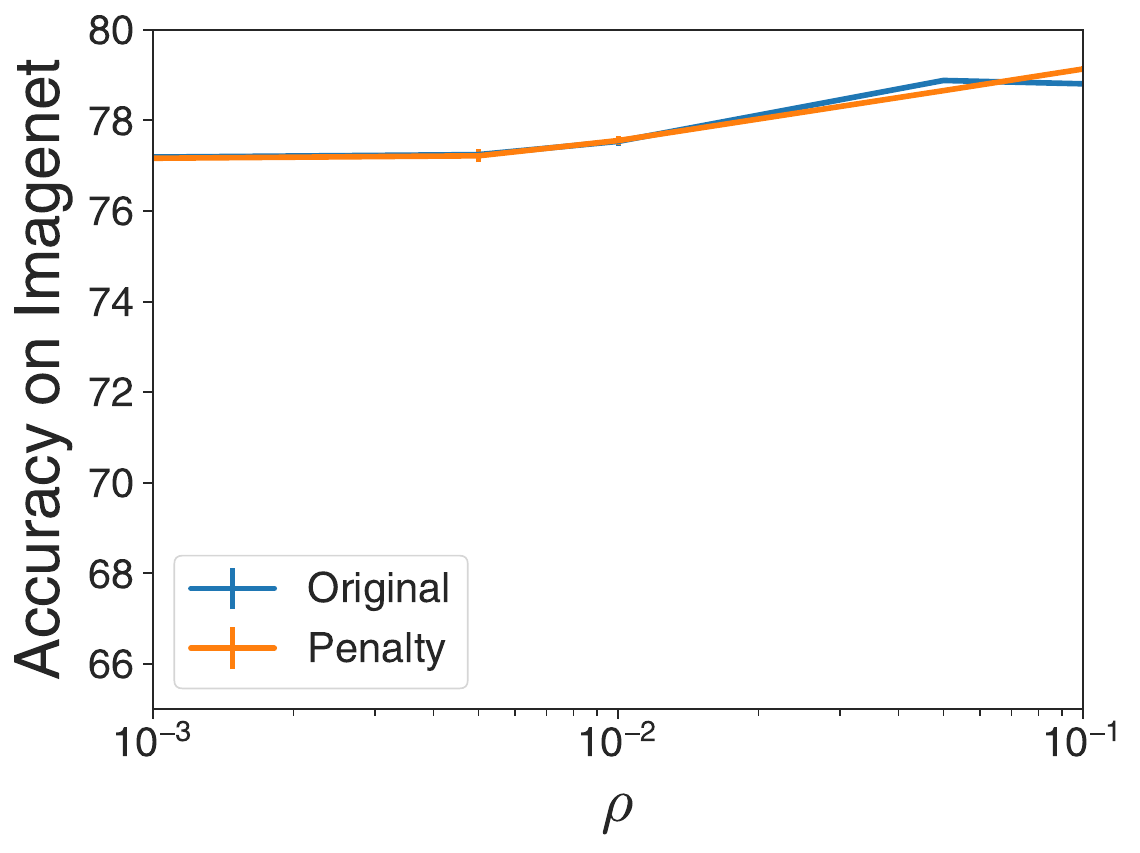}
         \caption{Imagenet with GELU}
         \label{fig:y equals x2}
     \end{subfigure}

        \caption{Test Accuracy as $\rad$ increases across different datasets and activation functions averaged over 2 seeds. For ReLU networks and large $\rad$, there is a significant difference between \texttt{P\texttt{SAM}} and \texttt{SAM}. \texttt{P\texttt{SAM}} with GELU networks more closely follows the behavior of \texttt{SAM}.}
        \label{fig:activations_rho_app}
\end{figure}

Figure \ref{fig:activations_rho_app} shows our experimental results comparing \texttt{P\texttt{SAM}} and \texttt{SAM} for Imagenet with different activation functions.
We already saw in Section \ref{sec:grad_pen} that 
\texttt{P\texttt{SAM}} behaves differently between the two activation functions;
by contrast, \texttt{SAM} is insensitive to them.
The original \texttt{SAM} algorithm implicitly captures the \mshort{} information
with the discrete $\rad$ step even in the ReLU case, while the gradient penalty,
which uses explicit Hessian-gradient products, does not.

This suggests that another way to combat poor NME performance is to incorporate
first order information from nearby points. There may be cases where this is more
efficient computationally; \texttt{SAM} requires $2$ gradient computations per step,
which is similar to the cost of an HVP.

\subsection{Penalty \texttt{SAM} vs. implicit regularization of SGD}

\label{app:pen_vs_step_size}

The analysis of \cite{smith2021origin} suggested that SGD with learning rate
$\lr$ is similar to gradient flow (GF) with \texttt{PUSAM} with $\rad = \lr/4$.
In this section we use a linear model to highlight some key differences between
\texttt{PUSAM} and the discrete effects from finite stepsize.

Consider a quadratic loss $\Lo(\th) = \frac{1}{2}\th^{\tpose}\Hmat\th$ for some
parameters $\th$ and PSD Hessian $\Hmat$. It is illustrative to consider gradient 
descent (GD) with learning rate $\lr$ and (unnormalized) penalty \texttt{SAM} with
radius $\rad$.

The gradient descent update rule is
\begin{equation}
\th_{t+1}-\th_{t} = -\lr(\Hmat+\rad\Hmat^2) \th_{t}
\end{equation}
The ``effective Hessian'' is given by $\Hmat+\rad\Hmat^2$
(see \cite{agarwala_sam_2023} for more analysis).
Solving the linear equation gives us
\begin{equation}
\th_{t} = \left(1-\lr(\Hmat+\rad\Hmat^2)\right)^{t} \th_{0}
\end{equation}
This dynamics is well described by the
eigenvalues of the effective Hessian - $\lambda+\rad\lambda^2$, where
$\lambda$ are the eigenvalues of $\Hmat$. The effect of the regularizer
is therefore to introduce eigenvalue-dependent modifications into the Hessian.

There is a special setting of $\rad$ which
can be derived from the calculations in \cite{smith2021origin}. Consider
$\rad = \lr/2$, and consider the dynamics after $2t$ steps. We have:
\begin{equation}
\th_{2t} = \left(1-\lr(\Hmat+\frac{1}{2}\lr \Hmat^2)\right)^{2t} \th_{0}
\end{equation}
which can be re-written as
\begin{equation}
\th_{2t} = \left(1-2\lr\Hmat+\lr^3\Hmat^3+\frac{1}{4}\lr^4\Hmat^4\right)^{t} \th_{0}
\end{equation}
To leading order in $\lr\Hmat$, this is the same as the dynamics for
learning rate $2\lr$, $\rad = 0$ after $t$ steps:
\begin{equation}
\th_{t} = \left(1-2\lr\Hmat\right)^{t} \th_{0}
\end{equation}

We note that these two are similar only if $\lr\Hmat\ll 1$. Under this
condition, $\lr\rad\Hmat^2 = \frac{1}{2}\lr^2\Hmat^2\ll \lr\Hmat$, and the
gradient penalty only has a small effect on the overall dynamics. In many
practical learning scenarios, including those involving \texttt{SAM}, $\lr\lambda$
can become $O(1)$ for many eigenvalues during training \cite{agarwala_sam_2023}.
In these scenarios there will be qualitative differences between using penalty
\texttt{SAM} and training with a different learning rate.

In addition, when $\rad$ is set arbitrarily, the dynamics of
$\lr$ and $2\lr$ will no longer match to second order in $\lr\Hmat$.
This provides further theoretical evidence that combining SGD with penalty
\texttt{SAM} is qualitatively and quantitatively different from training with a larger
learning rate.

\end{document}